\documentstyle[lingmacros,named]{article} 

\begin{document}
\input{psfig}
\bibliographystyle{named}  
\topmargin 0pt
\advance \topmargin by -\headheight
\advance \topmargin by -\headsep

\textheight 8.9in

\oddsidemargin 0pt
\evensidemargin \oddsidemargin
\marginparwidth 0.5in

\textwidth 6.5in

\setcounter{bottomnumber}{20}
\setcounter{topnumber}{20}
\renewcommand{\bottomfraction}{1}
\renewcommand{\topfraction}{1}
\setcounter{totalnumber}{30}
\renewcommand{\textfraction}{0}
\renewcommand{\floatpagefraction}{0}

\author{Marilyn A. Walker \\
Mitsubishi Electric Research Laboratories \\ 201 Broadway \\
Cambridge, Ma. 02139, USA \\{\tt walker@merl.com} }
\title{
The Effect of Resource Limits and Task Complexity on Collaborative Planning in
Dialogue \\
\begin{small} To Appear in Artificial Intelligence \end{small} }

\maketitle

\begin{abstract}

This paper shows how agents' choice in communicative action can be
designed to mitigate the effect of their resource limits in the
context of particular features of a collaborative planning task. I
first motivate a number of hypotheses about effective language
behavior based on a statistical analysis of a corpus of natural
collaborative planning dialogues. These hypotheses are then tested in
a dialogue testbed whose design is motivated by the corpus analysis.
Experiments in the testbed examine the interaction between (1) agents'
resource limits in attentional capacity and inferential capacity; (2)
agents' choice in communication; and (3) features of communicative
tasks that affect task difficulty such as inferential complexity,
degree of belief coordination required, and tolerance for errors.  The
results show that good algorithms for communication must be defined
relative to the agents' resource limits and the features of the task.
Algorithms that are inefficient for inferentially simple, low
coordination or fault-tolerant tasks are effective when tasks require
coordination or complex inferences, or are fault-intolerant.  The
results provide an explanation for the occurrence of utterances in
human dialogues that, prima facie, appear inefficient, and provide the
basis for the design of effective algorithms for communicative choice
for resource limited agents.

\end{abstract}

\section{Introduction}
\label{intro-sec}

Agents may engage in conversation for a range of reasons, e.g. to
acquire information, to establish a contract, to make a plan, or to be
social. At each point in a dialogue, agents must make communicative
choices about what to say and how and when to say it. This paper
focuses on agents' communicative choice in collaborative planning
dialogues, dialogues whose purpose is to discuss and agree on a plan
for future action, and potentially execute that plan. I will argue
that agents' choices in communicative action, their algorithms for
language behavior, must be determined with respect to two relatively
unexplored factors in models of collaborative planning dialogues: (1)
agents' resource limits, such as limits in attentional and inferential
capacity; and (2) features of collaborative planning tasks that affect
task difficulty, such as inferential complexity, the degree of belief
coordination required, and tolerance for errors.

A primary dimension of communicative choice is the degree of
explicitness. For example, consider a simple task of agent A and agent
B trying to agree on a plan for furnishing a two room house. Imagine
that A wants B to believe the proposition realized by \ex{1} and
believes that B can infer this from the propositions realized in
\ex{2}:\footnote{These examples are from the domain of Design-World to
be discussed in section \ref{dw-sec} and are abstractions from
naturally occurring examples in which the propositions realized here
are realized in a number of different ways. Here the focus is on the
logical relationships between the contents of each proposition:
\ex{2}a and \ex{2}b are minor premises and \ex{2}c is the major
premise for the inference under discussion.}

\enumsentence{ If we agree to put the green couch in the study, we
will have a matched-pair of furniture in the study.}

\eenumsentence{
\item I propose that we put the green couch in the study.  \item We
intend to put the green chair in the study.  \item Two furniture items
of the same color in the same room achieve a matched pair.  }

In naturally-occurring dialogues, A may produce utterances
realizing the propositions in \ex{1} to \ex{4}, or other variations
\cite{Sadock78,RCohen87,WJ82,Walker93c}.

\eenumsentence{
\item[A:] I propose that we put the green couch in the study.
\label{least-exp-examp}}

\eenumsentence{
\item[A:] We intend to put the green chair in the study.
I propose that we put the green couch in the study.
\label{more-exp-examp}}

\eenumsentence{
\item[A:]  Two furniture items of the same color in the
same room achieve a matched pair.
We intend to put the green chair in the study.
I propose that we put the green couch in the study.
\label{even-more-exp-examp}}

\eenumsentence{
\item[A:] I propose that we put the green couch in the study.
That will get us a matched pair.
\label{inf-exp-examp}}

The communicative choices in \ex{-3} through \ex{0} illustrate a
general fact: for any communicative act, the same effect can be
achieved with a range of acts at various levels of explicitness. This
raises a key issue: On what basis does A choose among the more or less
explicit versions of the proposal in 3 to 6?

The single constraint that has been suggested elsewhere in the
literature is the {\sc redundancy constraint}: A should not say
information that B already knows, or that B could infer. The {\sc
redundancy constraint} appears in the form of simple dictums such as
`Don't tell people what they already know', as Grice's Quantity Maxim
'do not make your contribution more informative than is
required'\cite{Grice75} and as constraints on planning operators for
the generation and recognition of communicative
plans\cite{AP80,Cawsey91,Cohen78,MooreParis93,LA90}. So, if we assume
that B knows \ex{-4}b and \ex{-4}c, then the only possibility for what
A can say is \ex{-3}.

The {\sc redundancy constraint} is based on the assumption that agent
A should leave implicit any information she believes that B already
knows or she believes that B could infer, in other words, that agent B
can always `fill in what is missing' by a combination of retrieving
facts from memory and making inferences. In section \ref{iru-sec}, I
will show that agents in natural dialogues consistently violate the
{\sc redundancy constraint}. I will argue that this should not be
particularly surprising since the {\sc redundancy constraint} is based
on several simplifying assumptions:

\begin{enumerate}
\item {\sc
unlimited working-memory assumption}: everything an agent knows is always
available for reasoning;
\item {\sc logical omniscience assumption}: agents are logically
omniscient;
\item {\sc fewest utterances assumption}: utterance
production is the only process that should be minimized;
\item {\sc no autonomy assumption}: assertions and proposals by Agent A are
accepted by default by Agent B.
\end{enumerate}

When agents are autonomous and resource-limited, these assumptions do
not always hold, and the problem of communicative choice remains.

The plan for the paper is as follows: section \ref{iru-sec} motivates
a number of hypotheses about the relationship of communicative choice,
resource limits and task features using evidence from natural
collaborative planning dialogues. These hypotheses are the basis of a
model of collaborative planning presented in section \ref{model-sec}.
Then section \ref{dw-sec} describes how the model is implemented in a
testbed for collaborative planning dialogues called Design-World,
which supports experiments on the interaction of agents' communicative
choice, resource limits, and features of the task. At this point, in
section \ref{method-sec}, I review the steps of the method applied so
far, and motivate the use of simulation as a method for testing the
hypotheses.  Section \ref{results-sec} presents the experimental
results and discusses the extent to which the hypotheses were
confirmed, and then section \ref{discussion-sec} discusses the
theoretical implications of these results and the extent to which they
can be generalized to other tasks, agent properties, and communication
strategies.

\section{Communicative choice in Dialogue}
\label{iru-sec}

Naturally occurring collaborative planning dialogues are design,
problem solving, diagnostic or advice-giving
dialogues\cite{Cawsey92a,RCohen87,Reichman85,Grosz77,WJ82,FJW86,Pollack86a,CS89,trains,WS88}.
In order to generate hypotheses about the relation of communicative
choice to agent properties and task features, this section examines
communicative choice in naturally occurring collaborative planning
dialogues. Most of the examples discussed below are excerpts from a
corpus of dialogues from a radio talk show for financial planning
advice \cite{PHW82},\footnote{The corpus consists of 55 dialogues from
5 hours of live radio broadcast, where each dialogue ranged in length
from 23 to 100 turns.} but I will also draw on data from collaborative
design, collaborative construction, and computer support dialogues
\cite{Cohen84a,WW90,WS88,WGR93}.

Dialogue, in general, is modeled as a process by which conversants add
to what is assumed to be already mutually believed or intended. This
set of assumed mutual beliefs and intentions is called the {\sc
discourse model}, or the {\sc common
ground}\cite{Webber78,Stalnaker78}.  In collaborative planning
dialogues, the conversants are attempting to add mutual beliefs about
the current state of the world and mutual beliefs and intentions about
a plan for future action to the discourse model.  It is obvious that
the efficacy of the final plan and the efficiency of the planning
process must be affected by agents' algorithms for communicative
choice.

However previous work has not systematically varied factors that
affect communicative choice, such as resource limits and task
complexity.  Furthermore, most previous work has been based on the
{\sc redundancy constraint}, and apparently, its concomitant
simplifying assumptions (but see
\cite{Zukerman,ZukermanMcConachy93,Lenke94}).

To explore the relation of communicative choice to effective
collaborative planning, the analysis of naturally occurring
collaborative planning dialogues in this paper focuses on
communicative acts that violate the {\sc redundancy constraint}. These
acts are {\sc informationally redundant utterances}, IRUs, defined
as:\footnote{The first part of the definition is a variation on
Hirschberg's definition of redundant\cite{Hirschberg85} which is used
in her theory of scalar implicature. This view of information is also
the basis of information theoretic work such as\cite{Barwise88b}.}

\begin{quote}
{\bf Definition of Informational Redundancy}  \\
\nopagebreak
An utterance $u_i$ is
{\sc informationally redundant} in a discourse situation $\cal S$

\begin{enumerate}
\item   if $u_i$ expresses a proposition $p_i$, and
another utterance $u_j$ that entails $p_i$ has already
been said in $\cal S$.
\item   if $u_i$ expresses a proposition $p_i$, and
another utterance $u_j$ that presupposes or implicates $p_i$ has already been
said in
$\cal S$.
\end{enumerate}
\end{quote}
\label{IRU-def}

A statistical analysis of the financial advice corpus showed that
about 12 \% of the utterances are IRUs.  As mentioned in section
\ref{intro-sec}, this should not be particularly surprising since
the definition of IRUs reflects several simplifying assumptions.  For
example, the definition reflects the {\sc logical omniscience
assumption} because it assumes that all the entailments of
propositions uttered in a discourse and certain default inferences
from propositions uttered in a discourse become part of the discourse
model.\footnote{Presuppositions and implicatures are two types of
default inferences\cite{Grice67,KP79,Gazdar79,Levinson83,Thomason90}.
The corpus analysis tagged defaults separately from entailments but
found no evidence for a functional difference (see Walker93c).} The
definition reflects the {\sc no autonomy assumption} because it
assumes that merely saying an utterance u$_j$ that expresses a
proposition p$_i$ is sufficient for adding p$_i$ to the discourse
model.  The fact that IRUs occur shows that the simplifying
assumptions are not valid.

The distributional analysis suggests that there are at least 3
functional categories of IRUs:

{\bf Communicative Functions of IRUs}: \\
\begin{itemize}
\item Attitude: to provide evidence
supporting beliefs about mutual understanding and acceptance
\item Attention: to manipulate the locus of attention of the
discourse participants by making a proposition salient.
\item Consequence: to augment the evidence supporting
beliefs that certain inferences are licensed
\end{itemize}

IRUs have {\sc antecedents} in the dialogue which are the utterances
that originally realized the content of the IRU either through direct
assertion or by an inferential chain; in the definition above $u_j$ is
an antecedent for $u_i$.  The 3 communicative functions of IRUs were
identified by correlations with distributional features based in part
on relations between the IRU and its antecedent, such as textual
distance, discourse structure relations, and logical relations.  The
distributional analysis also analyzed utterance features such as the
intonational realization of the IRU, the form of the IRU, and the
relation of the IRU to adjacent utterances.

Below, I will briefly give examples of each type of IRU.\footnote{Each
communicative function given above includes a number of subtypes that
will not be represented by these examples. In addition, the hypothesis
that IRUs are a rehearsal mechanism, i.e. agents repeat propositions
as an aid to memory, is tested in every experiment by the model of
Attention/Working memory.  The hypothesis that agents say IRUs because
they cannot think of anything else to say, (the {\sc dead air}
hypothesis), was considered in \cite{Walker92a}, but I as yet have
found no evidence to support it.  For example, other indications of
hesitation or planning what to say, such as disfluencies and long
pauses, are not associated with IRUs.} For each type I will explain
how the four simplifying assumptions of previous dialogue models
predict that the utterance is informationally redundant. Then we will
consider hypothetical agent and task properties under which IRUs
function as hypothesized above.

\subsection{Attitude IRUs \label{attitude-iru-sec}}

Attitude IRUs provide evidence supporting beliefs about mutual
understanding and acceptance by demonstrating the speaker's attitude
to an assertion or proposal made by another agent in dialogue. An
Attitude IRU, said with a falling intonation typical of a declarative
utterance, is given in \ex{1}-27, where M repeats what H has asserted
in
\ex{1}-26. M and H have been discussing how M and her husband can
handle funds invested in IRAs (Individual Retirement Accounts). In
\ex{1}, and in the other naturally occurring examples below, the {\sc
antecedents} of the IRUs are {\it italicized} and the IRUs are in
CAPS.

\eenumsentence{
\item[]
(24) H: That is correct.  It could be moved around so that each of you
have two thousand. \\ (25) M: I see. \\ (26) H: {\it Without penalty}.
\\ (27) M: WITHOUT PENALTY. \\ (28) H: Right. \\ (29) M: And the fact
that I have a, an account of my own from a couple of years ago, when I
was working, doesn't affect this at all.  }

The IRU in 27 provides direct evidence that M heard exactly what H
said \cite{CS89,Brennan90}. According to arguments elaborated below
and elsewhere \cite{Walker92a,Walker93c}, M's response indirectly
provides evidence that she accepts and therefore believes what H has
asserted.

\begin{figure}[htb]
\begin{center}
\rule{12cm}{.2mm}
\begin{tabular}{ll}
header: & {\sc inform}(speaker, hearer, proposition)\\
precondition:&{\sc know} (speaker,proposition) \\ want-precondition:&
speaker want {\sc inform}(speaker, hearer, proposition) \\ effect: &
{\sc know} (hearer,proposition)
\label{ap-inform-fig}
\end{tabular}

\rule{12cm}{.2mm}
\caption{Definition of the {\sc inform} plan operator in Allen and Perrault,
1980}
\end{center}
\end{figure}

The classification of \ex{0}-27 as an IRU follows from the {\sc
no-autonomy assumption}. The {\sc no-autonomy assumption} is usually
characterized as an agent being "co-operative" or "helpful". For
example, the motivation for the plan effect of the {\sc inform}
planning operator from \cite{AP80} in figure 1 is
that the hearer is cooperative. In other words, a cooperative hearer
always accepts and therefore believes (or knows) what the speaker has
previously asserted. But if the effect of the {\sc inform} act always
goes through, then there is no reason for M to choose to produce an
Attitude IRU in
\ex{0}-27, in response to H's inform in \ex{0}-26.

In recent work, the plan effect shown in figure 1 is treated as a
default \cite{JWW86,Reiter80,Perrault90,GS90}. Perrault's Belief
Transfer Rule handles inferring the default acceptance of assertions
(inform acts), while Grosz and Sidner's Conversational Default Rule
CDR2 handles inferring the default acceptance of proposals
\cite{GS90,Perrault90}.  In both cases, the default inference of
acceptance of an assertion of P or a proposal to achieve P depends on
the cooperativity of the hearer and on whether or not the hearer
previously believed or intended to achieve $\neg \rm{P}$.  However,
Attitude IRUs are produced in many situations, where there is no
reason for the default not to go through.  In advice giving dialogues,
the hearer is cooperative and the hearer does not previously believe
or intend to achieve $\neg \rm{P}$, yet Attitude IRUs are common when
the caller asks a the talk show host a question and then repeats or
paraphrases his response to the question with an Attitude IRU.

Clark and Schaefer proposed that Attitude IRUs provide positive
evidence of understanding \cite{CS87,CS89,BrennanHulteen95}. They
allow for understanding to be implicitly conveyed, but say that the
amount of explicit positive evidence should be `sufficient for current
purposes'.  However, Clark and Schaefer do not address the question of
belief transfer since they do not distinguish between indicating
understanding and indicating acceptance. Furthermore, they make no
predictions about what features of current purposes require more or
less positive evidence, and thus lead an agent to produce an Attitude
IRU.

Thus, neither the addition of defaults nor the positive evidence model
makes any predictions about when an agent should produce an Attitude
IRU, since the inference of acceptance goes through by default without
the Attitude IRU.

In order to explain the function of Attitude IRUs, the {\sc
no-autonomy assumption} must be replaced with the assumption that
hearers always either explicitly or implicitly accept or reject each
utterance act that is intended to change their beliefs or intentions.
In section \ref{model-sec}, these observations are incorporated into
ahe model of collaborative planning dialogue.  Results from testing
hypotheses related to the choice to produce Attitude IRUs are
presented elsewhere \cite{Walker92a,Walker93c,Walker94a}, and will not
be discussed further in this paper.

\subsection{Consequence IRUs}

Consequence IRUs make inferences explicit.  For example, consider
\ex{1}-17:

\eenumsentence
{\item[] (15) H: Oh no.  {\it I R A's were available as long as you
are not a participant in an existing pension} \\ (16) j. Oh I see.
Well I did work, {\it I do work for a company that has a pension} \\
(17) H: ahh.  THEN YOU'RE NOT ELIGIBLE FOR EIGHTY ONE \\
\label{elig-examp}}

In \ex{0}, \ex{0}-15 realizes a biconditional inference rule,
\ex{0}-16 instantiates one of the premises of this rule, and \ex{0}-17
realizes an inference that follows from \ex{0}-15 and \ex{0}-16, for
the particular tax year of 1981, by the inference rule of modus
tollens.

The definition of \ex{0}-17 as an IRU follows from the {\sc logical
omniscience assumption}.  If all entailments of utterances are
automatically added to the discourse model then \ex{0}-17 should not
occur.  However it is well known that neither human nor artificial
agents are logically omniscient
\cite{NormanBobrow75,Johnson-Laird91,Goldman86,Konolige86,HRT79}.
 Agents might not have enough time to make all the relevant inferences
even when they know all the relevant inference rules
\cite{Konolige85}, especially since producing and interpreting speech
in real time has heavy planning and processing requirements.  Thus
plausible hypotheses are that:

\begin{itemize}
\item HYPOTH-C1:  Agents choose to produce Consequence IRUs to
demonstrate that they made the inference that is made explicit.
\item HYPOTH-C2: Agents choose to produce Consequence IRUs to ensure
that
the  other
agent has access to inferrable information.
\end{itemize}

These hypotheses are motivated by the fact that agents are not
logically omniscient.  In addition, in the case of hypothesis C2,
agents may choose to produce Consequence IRUs to ensure that the other
agents have access to inferrable information in a timely manner, even
when, in principle, they believe the other agent is capable of making
the inference.

However, much of communicative efficiency relies on the fact that
agents do make some inferences from what has been said.  Thus
plausible refinements of hypotheses C1 and C2 are that:

\begin{itemize}
\item HYPOTH-C3:  The choice to produce a Consequence IRU is directly related
to a measure of `how hard' the inference is.
\item HYPOTH-C4: The choice to produce a Consequence IRU is  directly related
to a measure of `how important' the inference is.
\item HYPOTH-C5: The choice to produce a Consequence IRU  is directly related
to the degree to which the task requires agents to be coordinated on
the inferences that they have made.
\end{itemize}

Confirmation of these hypotheses entails that the {\sc fewest
utterances assumption} does not hold whenever processing effort is
relevant to achieving the conversational goals.

\subsection{Attention IRUs}
\label{attention-iru-sec}

Attention IRUs manipulate the locus of attention of the discourse
participants by making a proposition salient. Attention IRUs often
realize facts that are inferentially related to the assertion or
proposal that the speaker is making.  For example, consider
\ex{1} said by agent A to agent B while walking to work:

\eenumsentence
{\item Let's walk along Walnut St.
 \item IT'S SHORTER.
\label{walnut-examp}
}

Agent B already knew that the Walnut Street route was shorter, so, by
the {\sc redundancy constraint}, A should have simply said \ex{0}a.

The classification of \ex{0}b as an IRU reflects the {\sc unlimited
working memory} assumption.  If everything an agent knows is always
available for reasoning, then agents should never make communicative
choices to include utterances such as \ex{0}b.  However, it is well
known that human agents have limited attention/working memory \cite
{Miller56,NormanBobrow75,Baddeley86}, and resource bounded artificial
agents with limited time to access memory also have limited
attentional capacity.

If we define {\sc salient} propositions as those that are accessible
to a resource limited agent at a particular point in time
\cite{Prince81,Prince92}, then a possible hypothesis is that A said
\ex{0}b to provide B with a salient reason to accept A's proposal to
walk along Walnut St.  Similar observations apply to (\ex{1}):

\eenumsentence{
\item Clinton has to take a stand on abortion rights for poor women.
\item HE'S THE PRESIDENT.
\label{clinton-examp}
}

Here (\ex{0}b) is already known to the discourse participants, but
saying it makes it salient.  In order to account for \ex{0}b we must
modify the specific hypothesis above to reflect the difference in
utterance type between \ex{-1}a and \ex{0}a.  Utterance \ex{-1}a is a
{\sc proposal} whereas \ex{0}a is an {\sc inform}. In \ex{0}, A said
\ex{0}b to provide B with a salient reason to accept A's assertion
about Clinton's obligations.  Utterance \ex{-1}b is a {\sc warrant}
for adopting A's proposal in \ex{-1}a, and \ex{0}b is {\sc support}
for belief in A's assertion.\footnote{The relationship between these
utterances has been characterized as the inference of a discourse
relation \cite{MannThompson87}, or the inference of the speaker's
intention
\cite{MooreParis93}. Moser and Moore and Hobbs have argued
that these two views are functionally equivalent
\cite{MoserMoore93,Hobbs94}.}

\begin{itemize}
\item HYPOTH-A1:  Agents produce Attention IRUs to support the
processes of deliberating beliefs and intentions.
\end{itemize}

Hypothesis A1 means that the production of Attention IRUs is a surface
manifestation of the fact that agents' limited working memory limits
the accessibility of beliefs used as the basis of deliberation.  The
hypothesis that the function of Attention IRUs is to make information
salient to support interpretation and reasoning is formulated in the
{\sc discourse inference constraint}:

\begin{quote}
HYPOTH-A2: There is a {\sc discourse inference constraint}
whose effect is that inferences in dialogue are
derived from propositions that are currently discourse salient (in
working memory).
\end{quote}

The {\sc discourse inference constraint} is quite general since the
inferences that A intends B to make may be any inferences related to
the dialogue such as logical deductions, commonsense defaults,
inferring part of A's plan, or inferring relations such as {\sc
warrant} or {\sc support} \cite{WJ82,MooreParis93,Walker94c}.

A more complex example illustrating the relationship of limited
working memory, inferential processing, and agents' communicative
choice is dialogue \ex{1}. The caller E has been telling H, the
talk show host, how all her money is inveted and then poses a
question in \ex{0}-3:

\eenumsentence
{\item[] ( 3) E: ..... and I was wondering -- should I continue on
with the certificates or \\ ( 4) H: Well it's difficult to tell
because we're so far away from any of them. But I would suggest this
-- if {\it all of these are 6 month certificates and I presume they
are} \\ ( 5) E: {\it yes} \\ ( 6) H: {\it then I would like to see you
start spreading some of that money around} \\ ( 7) E: uh huh \\ ( 8) H:
Now in addition, how old are you? \\ . \\ (discussion about retirement
investments consisting of 14 utterances) \\ . \\ (21) E: uh huh and \\ (22a) H:
But as far as the
certificates are concerned, \\ (22b) I'D LIKE THEM SPREAD OUT A LITTLE BIT.
\\ (22c) THEY'RE ALL 6 MONTH CERTIFICATES. \\ (23) E: yes \\ (24) H: and I
don't like putting all my eggs in one basket.....
\label{certif-examp}
}

The utterances in 22b and 22c realize two propositions established as
mutually believed in utterances 4 to 7, thus they are IRUs. Utterance
8 initiates a subdialogue digression about retirement investments.
Since the discussion about retirement investments consist of 14
utterances in which the information in 4 to 7 is not discussed, a
plausible hypothesis is that, at 22a, H believes that the information
expressed in 4 to 7 is no longer salient \cite{Walker95b}.  However,
H expects E to use this information to make two inferences: (1) that
having all your investments in 6 month certificates is an {\sc
instance of} the negatively evaluated condition of having all your
eggs in one basket; and (2) that this is a {\sc warrant} for E to
adopt the intention to spread the certificates out a little bit.
Here, therefore, we see two types of inferences: a content-based
inference, {\sc instance of}, in the first case and a
deliberation-based inference, {\sc warrant}, in the second.  It appears
that H produces IRUs to ensure that these inferences get made and that
H is basing his communicative choice on the {\sc discourse inference
constraint}.

\begin{figure}[htb]
\begin{center}
\begin{tabular}{|r|c|c|}
\hline & & \\
& Consequence IRUs & Paraphrase IRUs   \\ \hline \hline &  &\\
Antecedents Salient  & 24 & 43\\
Antecedents Not Salient & 8 & 39    \\ \hline
\end{tabular}
\end{center}
\caption{Distribution of Consequence IRUs that make inferences explicit, as
compared with Paraphrases,
according to whether their antecedents are currently salient}
\label{conseq-sal-fig}
\end{figure}

In addition to the naturally occurring examples of Attention IRUs,
another source of evidence for the {\sc discourse inference
constraint} is the distribution of IRUs that make inferences explicit
such as the Consequence IRU in dialogue \ref{elig-examp}, \pageref{elig-examp}.
 Figure
\ref{conseq-sal-fig} contrasts the distribution of Consequence IRUs
and paraphrases, which are two different
ways in which an IRU can relate logically to the prior
discourse.\footnote{The other categories are repetitions, making
implicatures explicit and making presuppositions explicit.}
Paraphrases are syntactic or semantic transformations of a single
antecedent utterance \cite{McKeown82,Joshi64}. Inferences are
distinguished from paraphrases by requiring the application of a
logical inference rule such as modus ponens.  A key difference is that
inferences have multiple antecedents while paraphrases do not.  A
priori we would not expect inferences and paraphrases, as two types of
entailments, to distribute differently with respect to whether their
antecedents are salient.\footnote{For the corpus-analysis, salient
utterances are those within the last two turns. This measure is not
perfect but it  is replicable.} However figure
\ref{conseq-sal-fig} shows that {\sc inferences} are more likely to
have salient premises than {\sc paraphrases} ($\chi^2 = 4.835, p <
.05, df = 1$). This distributional fact provides evidence for the {\sc
discourse inference constraint} because whenever we have evidence that
an inference has been made, the premises are likely to be salient.

The data discussed above provide evidence for the {\sc discourse
inference constraint}, however it is clear that the effect of the
constraint is strongly determined by the limits on
working memory.   In
particular, a corollary of the constraint is that inferential
complexity can be directly related to the number of premises that must
be simultaneously salient for the inference to be made.
These hypotheses can be summarized as follows:

\begin{itemize}
\item HYPOTH-A3: The choice to produce an Attention IRU  is related
to   the degree  of inferential complexity of a task as measured by
the number of premises required to make task related inferences.
\item HYPOTH-A4: The choice to produce an Attention IRU  is related
to the degree to which an agent is resource limited in attentional
capacity.
\end{itemize}

Finally, it is obvious that various tasks can be characterized in
terms of the degree of inferential complexity, and that observations
about belief coordination similar to those made about Consequence IRUs
also apply to Attention IRUs, giving hypothesis A5.

\begin{itemize}
\item HYPOTH-A5: The choice to produce an Attention IRU  is related
to the degree to which the task requires agents to be coordinated on
the inferences that they have made.
\end{itemize}

In the next sections we will see how we can test these hypotheses.

\section{Modeling resource-limited collaborative planning dialogues}
\label{model-sec}

\begin{figure}[htb]
\centerline{\psfig{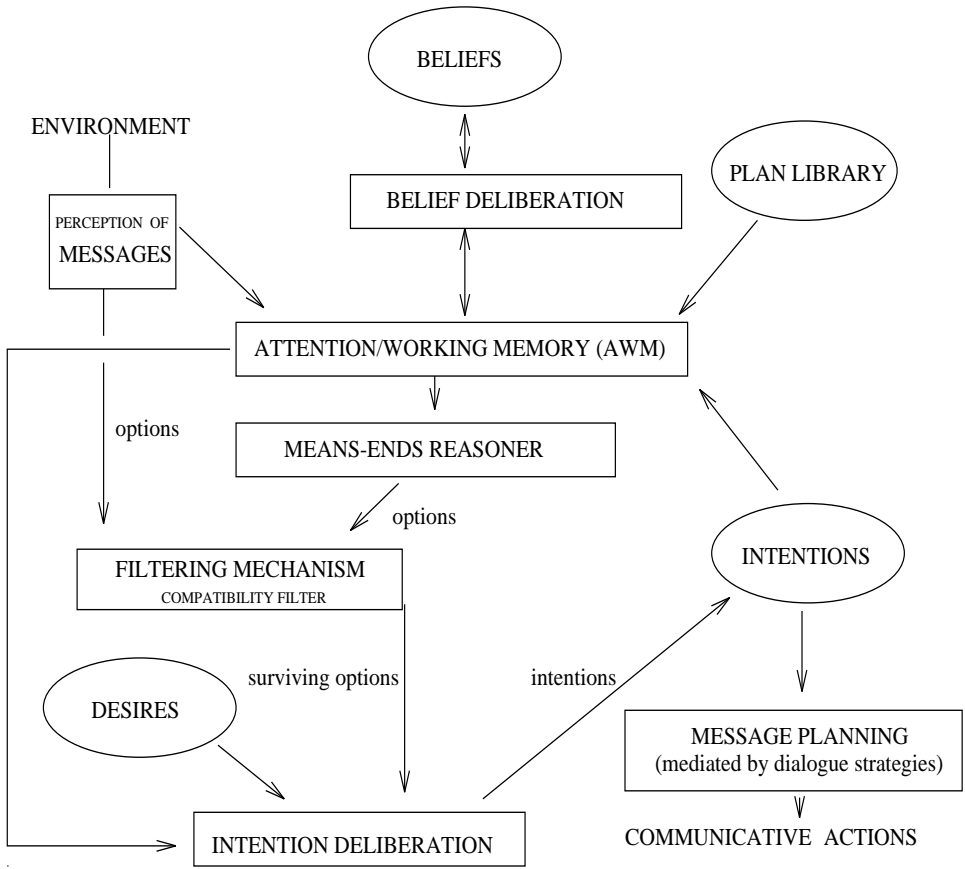}}
\caption{The IRMA Agent Architecture for
Resource-Bounded Agents with Limited Attention ({\sc awm})}
\label{irma-fig}
\end{figure}

The naturally occurring examples discussed in the previous section
gave rise to a number of hypotheses as to the situations in which
communicative choices to include IRUs could either improve the
efficacy of a collaborative plan or the efficiency of the dialogue by
which that plan was constructed.  In this section, I will specify the
details of a model of collaborative planning that will be used as the
basis of the dialogue simulation testbed in which the hypotheses can
be tested.  In thinking about models of collaborative planning, I have
found it useful to consider models in terms of 6 features:

\begin{enumerate}
\item agent architecture

\item role of resource limits: whether the agents constructing
the collaborative plan have limited resources,
 and thus whether there is an attempt to either maximize or minimize
any aspect of resource consumption, and if so which aspects.
\item  the mutual belief model:  whether  the function of the dialogue
is to establish mutual beliefs, and whether the mutual belief model is
binary
or  allows for defaults in mutual beliefs.
\item utterance act  types:  types of acts available for agents
to communicate with other agents and the effects of each act on
the cognitive state of the agents and the collaborative planning
process.
\item mixed-initiative: whether one agent is the initiator or both agents
have equal initiative.\footnote{This is also called `control'
\cite{WS88,Smith80}. Walker and Whittaker and Guinn argued that the
distribution
of control in natural dialogue is primarily determined by
whether information
relevant to the task is distributed between the agents or primarily
known by one agent \cite{WW90,Guinn93}.}
\item plan evaluation: how the   collaborative plan is evaluated and
what factors determine how good  the collaborative plan is.
\end{enumerate}

Most accounts of collaborative planning dialogues are not specific
about all of these features, although some accounts provide rich
models of particular features.  For example, Smith and Guinn provide a
richer model of mixed-initiative than that provided here
\cite{Guinn94,SmithHippBierman92}, and there are precise models of how
hearers infer the utterance act type or the intention underlying a
particular communicative action \cite{Allen83,Sidner85,LA90,Traum94}.
However, to my knowledge no previous work has included a specification
of the agent architecture, the relationship of the architecture to
language behavior, the role of resource limits, and the plan
evaluation process.  The remainder of this section provides a
specification for each of these features.

\subsection{Agent Architecture, Mutual Belief and Resource Limits}

Both the agent architecture and the role of resource limits are
addressed by adopting an agent architecture based on the IRMA
architecture for resource-bounded agents, shown in figure
\ref{irma-fig} \cite{BIP88,PollackRinguette90}.  The IRMA architecture
has not previously been used to model the behavior of agents in
dialogue.  The basic components of the modified IRMA architecture are:

\begin{itemize}
\item Beliefs: a database of an agent's beliefs. This includes beliefs that an
agent believes to be mutual to some degree.
\item Belief deliberation: decides what an agent wants to believe when
there is conflicting evidence.
\item Intentions: a database of an agent's intentions. This includes intentions
that an agent believes to be mutual to some degree.
\item Plan Library: what an agent knows about plans as recipes
to achieve goals.
\item Means-end reasoner: reasons about how to fill in existing
partial plans, proposing options that serve as subplans for the
plans an agent has in mind.
\item Filtering Mechanism: checks options for compatibility with
the agent's existing plans. Options deemed compatible are passed along
to the deliberation process.\footnote{The filtering mechanism
presented in
  \cite{BIP88} and used in Tileworld is more complex than that presented
here because that work explored the issue of when current intentions get
over-ridden.}
\item Desires: Agents may have different types of desires but here I assume
that their only desire is to maximize utility\cite{Doyle92}.
\item Intention Deliberation: decides which of a set of options
to pursue (by an evaluation based on desires such as maximizing utility).
\item Attention/Working memory ({\sc awm}): the limited attention
module constrains working memory and
the retrieval of current beliefs and intentions that are used by the
means-end reasoner.

\end{itemize}

For the purpose of modeling dialogue, the architecture has been
extended with a model of mutual belief that allows for different
degrees of mutual belief. For the purpose of exploring the effects of
resource-bounds on attention, this architecture has been extended with
a model of limited attention.  All of the modules are standard except
for the {\sc awm} module described in detail below, and the mutual belief
module which will be briefly described.

\paragraph{Attention/working memory model}

The model of limited {\sc awm} is a cognitively based model adapted
from \cite{Landauer75}, which fits many empirical results on human
memory and learning
\cite{Hellyer62,Landauer69,CollinsQuillian69,Sternberg67,Tulving67,AB73,Solomon92}.  The motivation for using a
cognitively based model of {\sc awm} is to model the behavior of
agents in naturally occurring dialogues and to test a theory of
collaborative communication with humans.\footnote{ Some of the
features of the model  hold for processors in general, such as
the feature that items that have been discussed more recently are more
likely to be accessible with little effort, and that incoming
information can displace other information from working memory.}

The key properties of the model are that (1) limits on {\sc awm} are a
parameter of the model and can be varied to explore different limits
\cite{Baddeley86}; (2) items encountered more recently are more likely
to be salient\cite{Landauer75}; (3) items encountered more frequently
are more likely to be salient \cite{HintzmannBlock71}.

These recency and frequency effects are a key aspect of the {\sc awm}
model for testing the hypothesized functions of IRUs. Below I will
discuss a particular implementation of this model and its role in
testing the hypotheses.

{\sc awm} is modelled as a three dimensional space in which propositions
acquired from perceiving the world are stored in chronological
sequence according to the location of a moving memory pointer.  The
sequence of memory loci used for storage constitutes a random walk
through memory with each loci a short distance from the previous one.
If items are encountered multiple times, they are stored multiple
times \cite{HintzmannBlock71}. The fact that the sequence of storage
locations is random means that the recency and frequency effects are
stochastically determined. This means that when this model is used in
simulation, the simulation produces different results each time.

When an agent retrieves items from memory, search starts from the
current pointer location and spreads out in a spherical fashion.  The
resource limited aspect of {\sc awm} follows from the fact that search is
restricted to a particular search radius defined in Hamming distance.
For example, if the current memory pointer loci is (0 0 0), the loci
distance 1 away would be (0 1 0) (0 -1 0) (0 0 1) (0 0 -1) (-1 0 0) (1
0 0). The actual locations are calculated modulo the memory size.

The limit on the search radius defines the subset of the belief and
intentions database that is {\sc salient}. In addition, the fact that
the pointer moves has the effect that the salient subset is always
changing.  Effectively, as new facts are added, others are displaced
and become no longer salient, so that the {\sc salient} predicate is
dynamic.

The search radius limit defines the {\sc awm} parameter that will be
varied in the experiments in section \ref{results-sec} in order to
test the effect of different resource limitations.  Experiments that
Landauer performed showed that, for a task requiring remembering
whether a word belonged to a list of words, the model can be
parameterized so that an {\sc awm} of 7 reproduces the human
performance results in \cite{Hellyer62}.  Since no systematic tests
have been performed for human performance on the collaborative
planning tasks investigated below, the experiments are run at {\sc
low, mid} and {\sc high awm} settings.  Human performance is assumed
to fall somewhere in the middle of these ranges.

\paragraph{Using {\sc awm} to implement the discourse inference constraint}
The model of Attention/Working Memory ({\sc awm}) provides a means of
testing the hypothesized {\sc discourse inference constraint}
introduced in section
\ref{attention-iru-sec}.  Remember that the discourse inference constraint
states that
inferences in discourse are derived from propositions that are
currently in working memory.  The {\sc awm} model, as shown in figure
\ref{irma-fig}, limits the beliefs accessible for means-end reasoning
and deliberation to the subset of beliefs that are currently in {\sc awm}
\cite{HRT79,Joshi78,JWW84,NormanBobrow75}.  These beliefs are defined
as being {\sc salient}.

This model contrasts with the standard view of inference, where if an
agent believes P and believes P$\rightarrow $Q, then the agent
believes Q.  The discourse inference constraint provides a principled
way of limiting inference in modeling humans by requiring the premises
of P and P$\rightarrow $Q to be salient. An axiomatization requires
the predicate {\sc salient} and inference rules as follows for each
inference rule schema\cite{Walker93c,Hobbs94}:

\begin{quote}
{\bf Inference under the Discourse Inference Constraint}:

Say(A,B,P) $\rightarrow$ Salient(B,P) \\

Salient(B,P) $\wedge$ BEL(B,P) $\wedge$ Salient(B,P$\rightarrow $Q)
 $\wedge$ BEL(B,P$\rightarrow $Q) \\ $\rightarrow$ BEL(B,Q)\\

\end{quote}

The first inference rule states that whenever agent A says an
utterance to B that realizes proposition P, that P becomes salient for
B. The second rule states that whenever a proposition P and an
inference rule, P$\rightarrow $Q, are both believed by B and salient
for B, then B can use them to infer Q. The model of {\sc awm} must be
consulted to determine when the {\sc salient} predicate holds.

\paragraph{Mutual belief model}
The model of mutual belief is based on Lewis's Shared Environment
model of mutual belief\cite{Lewis69,CM81,Barwise88b}, extended to
support different degrees of mutual belief by tagging beliefs with
qualitative endorsements at the time that they are formed and stored
in the beliefs database \cite{PCohen85,Gardenfors88,Galliers91b}.
Different degrees of mutual belief allow some actions to be left to
inference and some inferences to be defaults. This makes it possible
to distinguish between the explicit acceptance of a proposal and
the acceptance of a proposal inferred in the absence of
evidence to the contrary.   When agents are not logically
omniscient, it is possible to distinguish between mutual beliefs about what has
been mutually
inferred and information that has been discussed
in the dialogue. See \cite{Walker92a,Walker93c} for more detail.

\subsection{Discourse Acts, Utterance Acts, and Mixed Initiative}
\label{dial-sec}

\begin{figure}[htb]
\centerline{\psfig{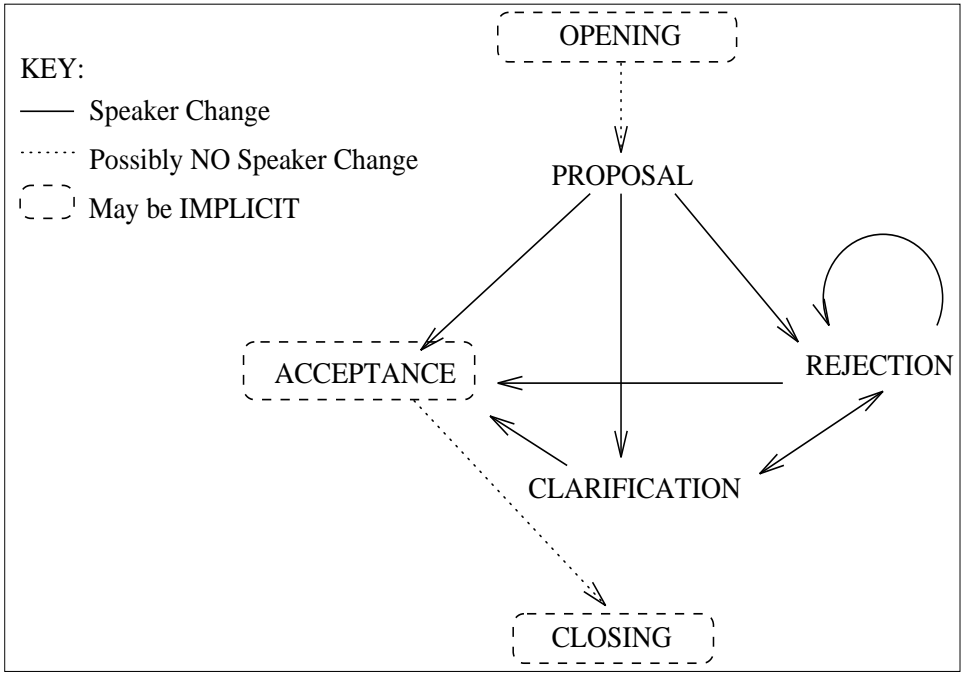}}
\caption{Finite State Model of Discourse Actions}
\label{dial-stat-fig}
\end{figure}

The overall structure of the discourse in collaborative planning
dialogues is primarily determined by the task structure
\cite{Power74,Grosz77,Litman85,Sibun91}.  Each subpart of the task
consists of a dialogue segment in which agents negotiate what they
should do for that part of the task.

As discussed in section \ref{attitude-iru-sec}, we wish to abandon the
{\sc no autonomy assumption}. The model should allow either agent to
initiate the dialogue or initiate a subdialogue about a new part of
the task \cite{WW90,Dahlback91}. For each agent to be able to do this,
knowledge about the task must be distributed between the participants
so that each participant has a basis for means-end reasoning and
deliberation.

 Furthermore, agents should be able to {\sc accept} or {\sc reject}
one another's proposals. Each plan step contributing to a higher level
goal must remain open for negotiation even if both agents are
committed to coming up with a collaborative plan for the higher level
goal.  This contrasts with models in which proposals for each plan
substep that the initiator makes must be accepted by the
non-initiator, once the non-initiator has agreed to work on a
collaborative plan
\cite{CohenLevesque91,GS90}.

\paragraph{Discourse  Acts and Mixed Initiative}
To engage in collaborative planning, agents take turns sending messages, and
each turn may consist of one or more {\sc discourse acts}.  Discourse
acts are {\sc opening, closing, proposal, acceptance, rejection} and
{\sc clarification}. These are higher level acts that are composed
of primitives called {\sc utterance acts}, which will be described
below.

 The schema of discourse actions shown in figure
\ref{dial-stat-fig} controls the sequence of discourse acts and which
discourse acts can be combined into a single turn.\footnote{This schema cannot
describe
all discourse action transitions in every type of dialogue
\cite{Levinson79,Levinson81,Schegloff87}. One required extension is to
allow multiple proposals to be simultaneously
open\cite{Rose95,Sidner94}.} The discourse act schema is the basis of
an algorithm by which agents achieve a {\sc collaborative-plan}.  For
each step in the domain plan:

\begin{enumerate} \item individual agents perform means-end reasoning
about options in the domain; \item individual agents deliberate about
which options are preferable; \item then one agent initiates a
subdialogue consisting minimally of a {\sc proposal} to the other
agent, based on the options identified in a reasoning cycle, about
actions that contribute to the satisfaction of their goals; \item then
the proposal may be subject to {\sc clarification}, after which it is
either {\sc accepted} or {\sc rejected} by the other agent, by
calculating whether it maximizes utility \end{enumerate}

This algorithm ties the discourse act schema in figure
\ref{dial-stat-fig} to the IRMA architecture. The requirement
that agents must indicate whether they accept or reject each proposal
follows from replacing the assumption of cooperativity in earlier work
\cite{AP80,GS90} with the {\sc collaborative principle}:

\begin{quote}
{\sc collaborative principle}:
Conversants must provide evidence of a detected discrepancy in belief as soon
as possible.
\end{quote}

The {\sc collaborative principle} was proposed in \cite{Walker92a},
and is an abstraction of the {\sc collaborative planning principles}
of Whittaker and Stenton (1988) and Walker and Whittaker (1990).  The
{\sc collaborative principle} means that speakers must monitor the
next action by the hearer in order to detect the effects of their
utterances. If the hearer continues the dialogue and provides no
evidence of a belief discrepancy, the inference of acceptance is
licensed as a default \cite{Galliers90,Walker92a}.

The fact that agents evaluate both assertions and
proposals before deciding what to believe or intend follows from the
IRMA agent architecture in figure \ref{irma-fig}.  As the figure
shows, incoming messages about intentions and beliefs are subject to
intention or belief deliberation.  This provides the basis for
abandoning the {\sc no autonomy assumption} while specifying {\bf why}
an agent would accept or reject another agent's proposal (see also
\cite{Galliers89,Galliers91a,Doyle92}). Agents evaluate assertions and
proposals from
other agents by assessing the support for assertions and the warrants
for proposals.  Finally, as figure \ref{irma-fig} shows, this
evaluation takes place under constraints of limited working memory,
since the beliefs that can serve as supports or warrants must be
salient for these processes to use them.

\paragraph{Utterance Acts}
Figure \ref{dial-stat-fig} shows the discourse act schema that
provides the basis for dialogue.  Discourse acts are composed of
utterance acts, which are the primitive acts that an agent can
actually perform.  Each discourse act can be performed in different
ways by varying the number and type of utterance acts that it consists
of.  For example, a proposal may or may not include additional
information that can convince the hearer, as in example
\ref{walnut-examp}.

There are seven utterance act types: {\sc open, close, propose,
accept, reject, ask} and {\sc say}, which are realized via the schemas
below:

\begin{quote}
 (propose ?speaker ?hearer ?option) \\ (Ask ?speaker ?hearer
?belief) \\ (say ?speaker ?hearer ?belief) \\ (Accept ?speaker ?hearer
?option) \\ (Reject ?speaker ?hearer ?belief) \\ (Reject ?speaker ?hearer
?option) \\ (Open ?speaker ?hearer ?option) \\ (Close ?speaker ?hearer
?intention) \\

\end{quote}

The content of each utterance act can be an {\sc options} and {\sc
intentions} representing a domain plan act constructed by means-end
reasoning and deliberation as shown in figure
\ref{irma-fig}.  An {\sc option} is an act that has not been committed
to by both agents. An {\sc intention} is an act that has been
committed to by both agents\cite{PollackRinguette90,BIP88}.  An option
only becomes an intention in the collaborative plan if it is {\sc
accepted} by both agents, either explicitly or implicitly.  The option
in a {\sc reject} schema is a counter-proposal and what is rejected is
the current proposal.

The content of an utterance act may also be a belief. These beliefs
are either those that an agent starts with, beliefs communicated by
the other agent, or inferences made by the agent during the
conversation. Beliefs in {\sc ask} actions have variables that the
addressee attempts to instantiate.  The belief in a rejection schema
is a belief that the speaker believes is a reason to reject the
proposal, such as a belief that the preconditions for the option in
the proposal do not hold
\cite{WW90}.

Examples of these utterance acts in dialogue will be given in section
\ref{dw-sec}.
Below how B processes each of the 7 messages that A can send is
specified. In the effects specified below, Store means store in {\sc
awm}, for eventual long-term storage in the beliefs database.  The
processing involved with each incoming message should be understood
with reference to the IRMA agent architecture.

\begin{enumerate}
\item Agent A: (propose ?speaker ?hearer ?option) \\
 Agent B: \begin{enumerate}
        \item Filter: Check whether ?option is compatible with current
beliefs, e.g. that no current beliefs contradict its preconditions.
 \item Infer and Store the preconditions of ?option
        \item Means-End Reason (ME-Reason) about Intention the ?option
              contributes to.
        \item Deliberate by evaluating the ?option against other options
              generated by Means-End reasoning.
        \item Indicate results of deliberation by an Accept or Reject.
 \end{enumerate}
\item  Agent A: (Ask ?speaker ?hearer ?belief); \\
 Agent B: retrieve beliefs matching ?belief
from Memory and respond with (say ?speaker ?hearer ?belief) with the
variable instantiated for each matching Belief.
\item Agent A: (say ?speaker ?hearer ?belief); \\
Agent B:  Store  ?belief
\item Agent A: (Accept ?speaker ?hearer ?option)\footnote{This is a
simplification since the form of the acceptance determines the
endorsement type on the mutual belief that is added to the beliefs
database.}\\
 Agent B: \begin{enumerate}
 \item Store (intend A B ?option)\footnote{This represents that both
agents are committed to the option while the binding of the ?option
specifies the agent who will execute the option.}
 \item Store (Act-Effects ?option)
 \end{enumerate}
\item Agent A: (Reject ?speaker ?hearer ?option) \\
 Agent B: \begin{enumerate}
 \item Deliberate ?option in comparison with own current proposal
that was rejected.
        \item Accept ?option if better than
              your current proposal.
        \item If rejecting ?option then
              reject with reason for rejection.
 \end{enumerate}
\item Agent A: (Reject ?speaker ?hearer ?belief); \\
 Agent B: Store  ?belief
\item Agent A: (Open ?speaker ?hearer ?option); \\
Agent B:  Mark the discourse segment that matches ?option as open
\item Agent A: (Close ?speaker ?hearer ?intention); \\
      Agent B: Close the  discourse segment
for ?intention
\end{enumerate}

These acts and their effects determine the structure of the dialogue
and its effect on the mental state of the conversants.

\subsection{Plan Evaluation}
\label{model-plan-eval-sec}

There are three components of the plan evaluation process that have
different effects on collaborative planning.  Two features are related
to the task definition and the third to the model
of evaluation applied:

\begin{enumerate}
\item  the degree of belief coordination: whether
some or all of the {\bf intentions} associated with a plan must be mutually
intended and whether any {\bf beliefs} related to the intended acts must
also be mutually believed;
\item  task determinacy and fault
tolerance: whether the task has only one solution, or is fault
tolerant or more or less satisfiable.
\item  the model must specify what is to
be optimized and whose resource consumption is to be minimized for
performance evaluation.
\end{enumerate}

Different theories of collaborative planning reflect different views
of the degree of belief coordination required for agents have a
collaborative plan.  The minimal approach is to not require the agents
to establish mutual beliefs at all \cite{Durfee,Guinn94}. Rather
agents divide up the plan into subcomponents and separately plan each
component, without requiring agreement on how the subcomponents are
planned.  At the next level of belief coordination, it is common to
require the intended acts of the collaborative plan to be mutually
intended \cite{GS90,Traum94,LCN90,Thomason90b}.  At the highest level
of belief coordination, the agents must both mutually intend all
intentions and mutually believe any beliefs that support the plan such
as the {\sc warrant} beliefs that provide reasons for adopting a step
of the plan.  In addition, it is possible to require that inferences
about other goals that the intended actions will satisfy should also
be mutually believed.

In this work, the assumption is that the degree of belief coordination
required is a feature of the task.  The minimal level in the
experiments discussed below will be that all intentions must be
mutually intended, and the experiments will vary whether {\sc
warrants}, and inferred intentions that are derived from explicitly
discussed intentions.

Task determinacy and fault tolerance have an effect on communicative
choice in collaborative planning because they are directly related to how
much uncertainty is tolerable in the plan.  If a partial plan has some
utility, then making a mistake or only constructing a partial plan is
not catastrophic.
For task determinacy, I assume that a measure of the quality of the
final plan can be determined from the utility of each step in the
plan, and that partial plans can also be evaluated, so that the task
is more or less satisfiable.

With respect to evaluating performance, I assume that the agents in a
collaborative planning dialogue are working as a team, and as a team
they attempt to optimize the team's performance and minimize the
team's consumption of resources.  This follows from Clark's assumption
that conversants in dialogue attempt to achieve their dialogue
purpose with {\sc least collaborative effort}\cite{CS89,CW86,CB90}.
This approach contrasts with other approaches in which agents only
participate in communication to the degree that it maximizes their own
expected utility \cite{Durfee}.

A final choice has to do with which processes collaborative effort
consists of.  A common assumption is that the number of utterances is
the primary efficiency measure \cite{Grice75,Chapanis75}; this is the
{\sc fewest utterances assumption}. Since all types of IRUs violate
this assumption, in this work collaborative effort is defined with reference to
the agent
architecture and to all the processes required in collaborative
planning, i.e. (1) retrieval processes necessary to access previously
stored beliefs in memory; (2) communicative processes related to
generating and interpreting utterances; and (3) reasoning processes
that operate on beliefs stored in memory and those communicated by
other agents. With respect to the IRMA architecture (figure
\ref{irma-fig}), retrieval processes are those that access {\sc awm},
the plan library and the beliefs and intentions databases,
communicative processes are the modules for perception and generation
of messages, and inferences are the combined processes of
deliberation, means-ends reasoning, and filtering.  Collaborative
effort includes the costs for both agents for all of these processes:

\begin{quote}
\begin{tabular}{l}
{\sc collaborative effort} $=$  \\
(the total cost of communication for both agents)\\
+ (the total cost of inferences for both agents)\\
+ (the total cost of retrievals for both agents)
\end{tabular}
\end{quote}

Collaborative effort is defined for the whole dialogue and not on a
per utterance basis. This definition and the other assumptions support the
specification of
the plan evaluation process.  Given the above definitions, performance
is the difference between a measure of the quality of the problem
solution and {\sc collaborative effort}.

\begin{quote}
{\sc performance} $=$
  {\sc quality of solution} -- {\sc collaborative effort}
\end{quote}

Since the agents' desires are  simply to maximize
utility, the quality of the solution is measured by the utility
of the resulting plan with respect to the agents' utility functions.

\section{Design-World}
\label{dw-sec}

\subsection{Methodological Basis of Design-World}
\label{method-sec}

Design-World is a testbed for a theory of collaborative communication,
which instantiates the model of collaborative planning in dialogue
discussed in section \ref{model-sec}.  In order to motivate the use of
the Design-World testbed in developing and testing a defeasible theory
of communication in collaborative planning, this section first
describes Design-World as an instance of a general method, and then
describes the testbed and its implementation as well as the task and
communication parameters. The method can be characterized by the steps
below:\nocite{Cohen91}

\begin{enumerate}
\item Generate hypotheses about the features of a model of collaborative
planning dialogues from a statistical analysis of human-human
dialogue corpora.
\item Produce a functional characterization of the model, specifically
including the parameters that could affect task outcome, or claims
about the efficacy of the model.
\item Implement the model  as a testbed so that (some of ) these
parameters can be controlled, while using independently motivated modules for
other aspects of testbed.
\item Test the hypotheses and the resulting  model  against
different situations controlled by parameter settings.
\end{enumerate}

The hypotheses that were generated by the statistical analysis of the
dialogue corpora were discussed in section \ref{iru-sec}. These
hypotheses are roughly that under constraints of resource bounds, task
inferential complexity, task fault tolerance, and task requirements
for belief coordination, communicative choices to include IRUs can
reduce collaborative effort or increase the quality of solution of the
collaborative plan.

The next step is to produce a functional characterization of the model
(a partial formalization). In section \ref{model-sec}, I discussed
features of a model of collaborative planning that interact with an
agent's autonomy, resource limits and communicative choices.  In
nonexperimental work, the model is the final result of the research.
However, this leaves the model and the claims that motivated the model
empirically unverified. In formal characterizations, many simplifying
assumptions need to be made, and it is not always clear that the
results carry over to complex domains where the simplifying
assumptions do not hold. While the model presented here is
empirically based on statistical analysis of a corpus of naturally
occurring dialogues, many of the hypotheses discussed above are
related to models of agents' processing.  Corpus analysis can only
provide weak support for these hypotheses.  Thus another source of
empirical verification is desirable in order to develop a
well-specified and defeasible theory.  This is the motivation for the
Design-World testbed.

Next, it is necessary to consider the parameters that could affect the
outcome or claims about the efficacy of the model and then implement
the model as a testbed so that at least some of these parameters can
be controlled. The use of independently motivated modules for other
aspects of the testbed guarantees that the testbed actually tests
something, and also makes it less likely that the testbed is a case of
`experimentation in the small' \cite{HPC93}.

Parameters that affect the efficacious use of IRUs have already been
discussed: these include resource bounds, task inferential complexity,
and requirements for belief coordination.  The {\sc awm} model
introduces a parameter for resource bounds, and is implemented as part
of the IRMA architecture for resource limited agents, which is
independently motivated \cite{BIP88,PollackRinguette90}. The {\sc awm}
model itself is also independently motivated, since it reproduces in
simulation many well known results on human memory and learning
\cite{Landauer75,Solomon92}. In addition, the utterance acts and their
effects are independently motivated by other research on collaborative
planning dialogues and by the statistical analysis of dialogue corpora
\cite{WS88,WW90,Sidner94,Stein93,Maier95,Carletta92,ChuCarberry95,Traum94}.

To introduce parameters related to tasks, the testbed is designed
around a simple task where two agents must form a collaborative plan
as to how to arrange some furniture in the rooms of a two room
house. The task is based on cooperative design tasks used
for experiments on distributed cooperative work for which a corpus of
dialogues was available \cite{WGR93}.
However, the simple task can be varied along three dimensions: (1)
inferential complexity; (2) degree of belief coordination required;
(3) tolerance for errors and usefulness of partial solutions.

These three task dimensions represent very different tasks.  For
example, varying the task by increasing inferential complexity
provides information on the performance of agent communication
algorithms in simple versus inferentially complex tasks.  The
dimensions enable us to generalize from the specific task in
Design-World to real world tasks.  Section \ref{domain-sec}
will introduce the Standard version of the task, and then Section
\ref{task-def-sec} will describe the task variations.

To introduce parameters related to the interaction of communicative
choice with task complexity and resource limits, agents are designed
so that they vary their communication strategies to include or not
include IRUs.  Section \ref{comm-choice-sec} will describe the
communicative choice parameters that will be used in the experimental
results presented in section \ref{results-sec}.

The experiments reported in section \ref{results-sec} will examine the
interaction of three factors: (1) resource limits; (2) communicative
strategies; and (3) task definition.  The experiments in the testbed
have several functions: (1) they demonstrate that the model can be
implemented; (2) they highlight potential flaws in the model; and (3)
they provide empirical verification of hypotheses about the function
of particular communicative strategies beyond that provided by corpus
analysis and researcher's intuitions.  This section describes the
domain, the implementation of the collaborative planning model in this
domain, the communicative strategies and the task variations.

\subsection{Design World Collaborative Planning Domain}
\label{domain-sec}

\begin{figure}[ht]
\centerline{\psfig{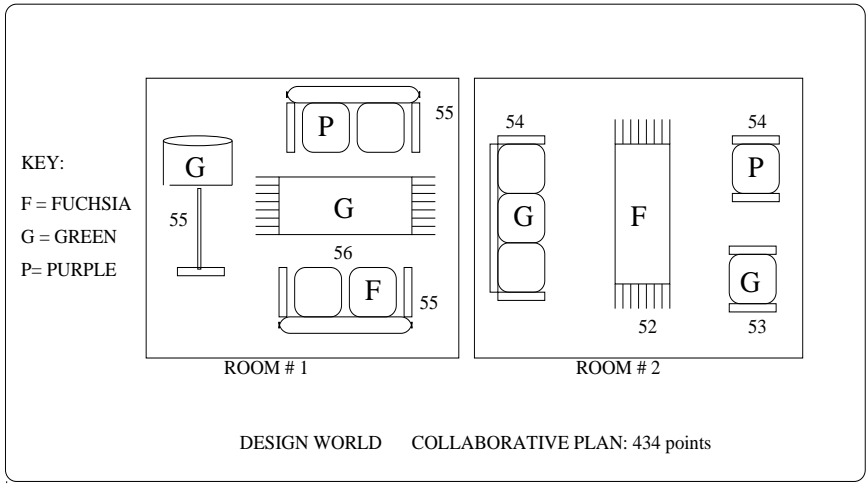}}
\caption{One Final State for Design-World Standard Task: Represents the
Collaborative Plan Achieved by the Dialogue, 434 points}
\label{final-state-fig}
\end{figure}

In Design-World, two artificial parametrizable agents converse in
order to agree on the design of the floor plan of a two room house
\cite{WGR93,Walker93c}.  The {\sc design-house} plan requires the
agents to agree on how to {\sc design-room-1} and {\sc design-room-2}.
At the beginning of the simulation, both agents know the structure of
the {\sc design-house} plan and that it requires four furniture items
in each room. Each agent has 12 items of furniture that can be used in
the plan.  A potential final collaborative design plan negotiated via
a (simulated) dialogue is shown in Figure \ref{final-state-fig}.

Each furniture item has a value that is used both as a {\sc warrant}
to support deliberation (the utility of an act that is under
consideration), and as the basis for calculating the {\sc
quality of solution} of the final plan, as discussed in section
\ref{model-sec}. The values on the furniture items range from
10 to 56, and both agents' furniture items range over these values.
Since beliefs about these values will be used to test hypotheses about
the function of IRUs, both agents know the values of all the furniture
items at the beginning of the dialogue.

Design-World agents negotiate a collaborative plan following the
collaborative planning algorithm discussed in section \ref{model-sec}.
Each agent carries out means-end reasoning about the furniture items
that can be used in the design.  Means-end reasoning generates {\sc
options} -- these options are the content of {\sc proposals} to the
other agent to {\sc put} a furniture item into a room.  This is
illustrated by Dialogue \ex{1}, which includes the artificial language
that the agents communicate with and a gloss automatically generated
from that language in {\it italics} for part of designing room-1.

\eenumsentence{

\item[1:]
BILL: {\it  First, put the green rug in the study.} \\
(propose agent-bill agent-kim option-10: put-act (agent-bill green rug room-1))

\item[2:]
KIM: {\it Then, let's put the green lamp in the study.} \\
(propose agent-kim agent-bill option-33: put-act (agent-kim green lamp room-1))

\item[3:]
BILL: {\it Then, let's put the green couch in the study.} \\
(propose agent-bill agent-kim option-45: put-act (agent-bill green couch
room-1))

\item[4:]
KIM: {\it No, instead let's put the purple couch in the study.} \\
(reject agent-kim agent-bill option-56: put-act (agent-kim purple couch
room-1))

\item[5:]
BILL: {\sc  Putting in the green couch is worth  56} \\
(say agent-bill agent-kim bel-265: score  (option-45:  put-act
(agent-bill green couch room-1) 56))

\item[6:]
BILL: {\it It is better to put the green couch in the study.} \\
(reject agent-bill agent-kim option-56: put-act (agent-bill green couch
room-1))

\label{bill-kim-dial-examp}
}

At the beginning of the dialogue, Agent-Kim has stored in memory the
proposition that (score green-rug 56). When she receives Bill's
proposal as shown in (\ex{0}-1), she evaluates that proposal in order
to decide whether to accept or reject it. As part of evaluating the
proposal she will attempt to retrieve the score proposition stored
earlier in memory.  Thus the propositions about the scores of
furniture items are {\sc warrants} for supporting deliberation.

As discussed in section \ref{model-sec}, the agents retain their
autonomy even though the agents both want to agree on a plan for
designing the house.  Thus, on receiving a proposal, an agent
deliberates whether to {\sc accept} or {\sc reject} it
\cite{Doyle92,Walker94a}.  Proposals 1 and 2 are inferred to be implicitly {\sc
accepted} because
they are not rejected \cite{WS88,WW90}. This follows from the
{\sc collaborative principle} discussed in section
\ref{model-sec}.  If a proposal is {\sc accepted},
either implicitly or explicitly, then the option that was the content
of the proposal becomes a mutual intention that contributes to the
final design plan \cite{Power84,Walker92a,Sidner94}.

Agents {\sc reject} a proposal if deliberation leads them to believe
that they know of a better option, based on evaluating the utility of
the competing options they have generated by means-end reasoning.  For
example, in (\ex{0}-4) Kim rejects the proposal in (\ex{0}-3), for
pursuing option-45, and proposes option-56 instead.  The form of the
rejection as a counter-proposal is based on observations about how
rejection is communicated in naturally-occurring dialogue as codified
in the {\sc collaborative planning principles} \cite{WW90}. When an
agent intends to reject another agent's rejection, as in \ex{0}-5 and
6, the agent includes additional information to support its proposal.
In \ex{0}-5, agent-bill reminds agent-kim of the value of the green
couch, before rejecting agent-kim's proposal.

\subsection{Agent architecture implementation in Design-World}
\label{irma-sec}

The agent architecture used in the Design-World simulation environment
is the modified IRMA architecture, shown in figure \ref{irma-fig} and
discussed in section \ref{model-sec} \cite{BIP88,PollackRinguette90}.
The only aspects of the architecture that are specific to Design-World
are the plan library, the way that {\sc awm} is implemented, and the way
that belief deliberation is implemented.

For the experiments below, the total size of {\sc awm} is set to 16, but
memory is wrap-around, and there is no overwriting.  If the path of
the memory pointer retraces its steps so that the current memory loci
already has something stored in it, the new item is simply added. Thus
memory capacity is unbounded.

Since hypothesis A4 relates to the degree to which {\sc awm} is limited, we
want to be able to compare the performance of agents who are more or
less attention limited. Thus, all experiments make comparisons between
different communicative strategies over three ranges of of {\sc awm}
settings; the {\sc awm} search radius parameter varies from {\sc low awm}
(radius of 3 and 4), to {\sc mid awm} (radius of 6 and 7) to {\sc high
awm} (radius of 11 and 16). {\sc low awm} agents are severely
attention limited agents, wherease almost everything an agent knows is
salient for {\sc high awm} agents.

The limits on {\sc awm} plays a critical role in determining agents'
performance. Remember that only {\em salient} beliefs can be used in
means-end reasoning and deliberation, so that if the warrant for a
proposal is not salient, the agent cannot properly evaluate a
proposal. However, if the agent only knows of one option, the agent
can accept the proposal on the assumption that any option is better
than doing nothing.  Section
\ref{results-sec} will show the impact of resource limits on
performance.  For more detail see \cite{Landauer75,Walker94a}.

The implementation of belief deliberation, for the purpose of
Design-World was tied directly to Landauer's {\sc awm} model. In that model,
nothing stored in memory is ever deleted or modified. Rather, new
beliefs are added which effectively compete with beliefs that are
already present. Thus an agent's belief deliberation process depends
on collecting a set of related beliefs which may be contradictory, and
applying an algorithm to determine what the agent believes
(see\cite{Galliers90,Gardenfors90}). The fact that new
information about the the state of the world supercedes old
information is an emergent property of the belief retrieval mechanism:
beliefs recently added are more likely to be retrieved. However the
stochastic aspect of retrieval means that it is possible for an agent
to decide to believe ``out-of-date'' propositions, and ``forget''
recent changes in the world. As we will see in section
\ref{results-sec}, this means that, in cases where the outdated beliefs
were encountered with greater frequency and thus stored in memory
repeatedly, that agents who can access all of their beliefs are more
likely to decide to believe out of date beliefs.\footnote{It is
unclear whether this prediction of the belief deliberation algorithm
is consistent with human performance.
However it is easy to think of examples of humans making the kind of
error that this model would predict.  For example, I commonly believe
(falsely) that I have eggs at home in the refrigerator, even though I
used them the previous evening for quiche.}

The plan library contains domain plans for Design-House and its
subgoals, as well as discourse plans for the discourse acts shown in
figure \ref{dial-stat-fig}. The discourse plans will be discussed
in detail in section \ref{comm-choice-sec}.

\subsection{Design-World Tasks}
\label{task-def-sec}

\begin{figure}[htb]
\centerline{\psfig{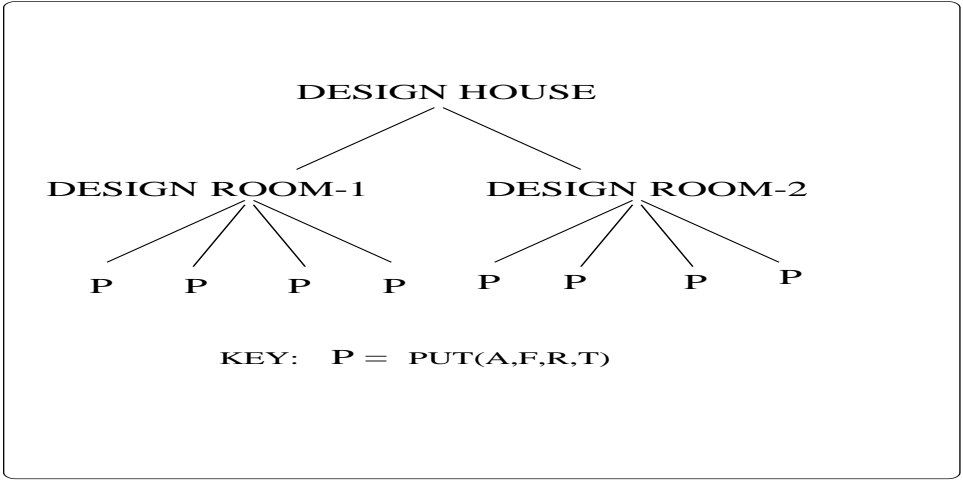}}
\caption{Standard Version of the Task, Fault tolerant
and Partial Solutions acceptable.}
\label{standard-fig}
\end{figure}

The Design-World task as a plan is simple since it involves linear
planning of subgoals which contribute to higher level goals, as shown
in figure \ref{standard-fig}. However, the task is easily modified
according to the general task features discussed above so that it is
more difficult to perform well. These modifications are applicable to
other tasks besides the testbed task, and affect the degree to which
different aspects of the task contribute to the performance
evaluation.

There are 4 versions of the task that will be used to test the
hypotheses introduced in section \ref{iru-sec}: Standard,
Zero-Nonmatching-Beliefs, Matched-Pair, and Zero-Invalid.  The
Standard task is  inferentially simple, fault tolerant
and requires low levels of belief coordination. The other tasks are
more difficult because they increase the degree of belief coordination
required, and magnify the effect of mistakes.  The
Zero-Nonmatching-Beliefs and Matched-Pairs tasks each explore
different aspects of belief coordination and inferential complexity.
The Zero-Invalids task is fault intolerant.

\paragraph{Standard Task}

The Standard task  provides a baseline and is
inferentially simple, fault tolerant and requires low levels of belief
co-ordination.  The Standard task is defined so that the {\sc quality
of solution} for a particular dialogue consists of the sum of all the
furniture pieces for each valid step in the plan. In addition, the
task is defined so that partial solutions are possible. Any number of
furniture items in a room is a valid plan, rather than requiring that
each room {\bf must} have all four furniture items. This choice about
task determinacy makes it possible to see the gradient effect on
performance of different resource restrictions.

In addition the
Standard task is fault tolerant.  If agents make a mistake in planning
and insert invalid steps in their collaborative plan, the point values
for invalid steps in the plan are simply subtracted from the score.
Thus in the Standard task agents are not heavily penalized for making
mistakes due to inserting steps in plans that are not actually
executable.

The Standard task is inferentially simple because the agent's only
inferences are those by means-end reasoning to generate options, those
by deliberation to evaluate options, and act-effect inferences after
committing to an action. Each of these inferences rely on one premise:
the premise (has ?agent ?item) supports means-end reasoning, the
premise (score ?item ?score) supports deliberation, and the premise
(intend A B ?option) supports inferring the effect of ?option.  Thus,
none of these processes require multiple premises to be simultaneously
salient.  However, it is possible to test hypotheses about processing
effort in the Standard task by making it easier to access these
inferential premises. It is also possible to test the effect of
resource limits since these premises must be accessible to perform
optimally on the task.

The degree of belief coordination in the Standard task is low because
agents are only required to coordinate on the intentions corresponding
to put-acts as shown in figure \ref{standard-fig}. These intentions
are always explicitly discussed so that coordination is always
achieved.

\begin{figure}[htb]
\centerline{\psfig{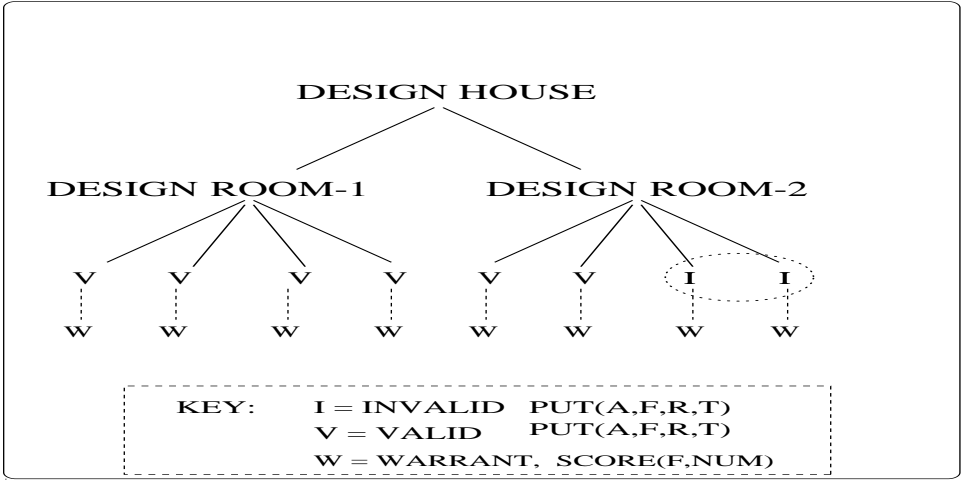}}
\caption{Tasks can differ as to the level of mutual belief required.
Some tasks require that W, a reason for
doing P, is mutually believed and others don't.}
\label{task-eval-fig}
\end{figure}

\paragraph{The Zero-Nonmatching-Beliefs task}
The Zero-Nonmatching-Beliefs task  increases the degree of
belief coordination by requiring agents to base their deliberation
process on the same beliefs.  They must have the same {\sc warrants}
for adopting an intention in order to do well on this task. Figure
\ref{task-eval-fig} shows the structure of beliefs about intentions
and warrants for the Design-House goal. In the
Zero-Nonmatching-Beliefs task, as shown, the warrants underlying
intentions must also be mutually believed.  This is not generally
required in forming a collaborative plan because agents A and B can
mutually believe that they have maximized utility without necessarily
agreeing on what that utility is. Furthermore, in the general case,
when agents have only one option under consideration, they do not need
to evaluate the utility of that one option in order to decide whether
to accept or reject it.

The Zero-Nonmatching-Beliefs task provides a basis for testing
hypotheses A1 and A5, introduced in section \ref{iru-sec} by increasing
the degree of belief coordination required to perform well on the
task, where the beliefs are those used in deliberation.

The Zero-Nonmatching-Beliefs task models particular types of
real-world tasks since it is not always necessary for agents to agree
on the reasons for carrying out a particular action.  For example, in
the negotiation between the union and the management of a company, any
agreement that is reached is agreed to by each party for different
reasons. An agreement for a shorter work week is supported by the
union because more overtime pay is possible for those who want to work
more and is supported by the management because the company's
insurance premiums will be lower. However, if two agents agree on a
plan, but have different reasons for doing so, they may change their
beliefs and their intentions under different conditions. The most
stable, long-term, collaborative plans will be those in which agents
agree on both the actions to be performed, as well as the reasons for
doing those actions. Under these conditions the agents will be more
likely to revise their intentions in a compatible way and intention
revision should be simpler.  Thus the Zero-Nonmatching-Beliefs task
examines one extreme of belief coordination for deliberation.

\begin{figure}[htb]
\centerline{\psfig{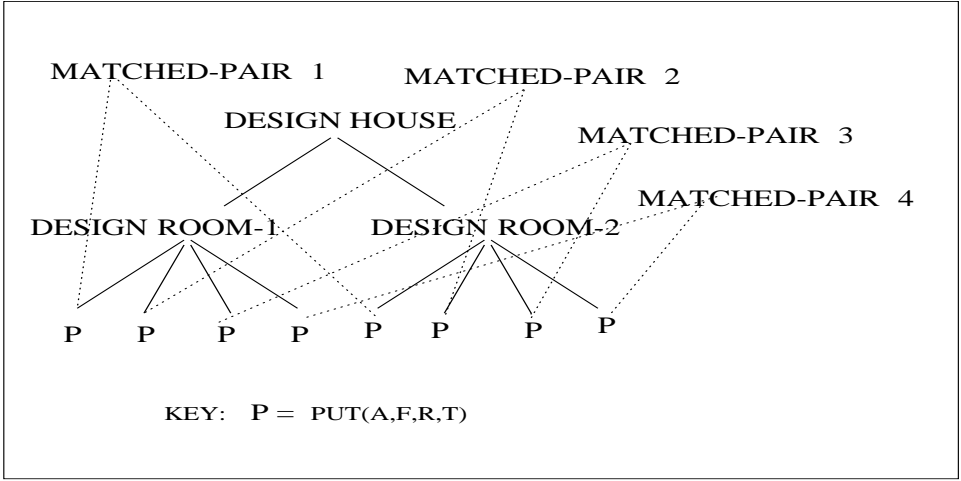}}
\caption{Making additional Inferences: Matched Pair Two Room Task.
Each PUT intention contributes both to a Design-Room goal as well
as a Matched pair goal}
\label{mp-task-fig}
\end{figure}

\paragraph{Matched Pair Tasks} Another aspect of belief coordination has to do
with coordinating
beliefs based on inferences.  There are two task definitions that
increase inferential complexity by increasing the number of
independent premises that must be simultaneously available in working
memory. These are: (1) Matched Pair Same Room, and (2) Matched Pair
Two Room. Figure \ref{mp-task-fig} shows the Matched Pair Two Room
version of the task.  Each intention to {\sc put} a furniture item in
a room can potentially contribute to another intention of achieving a
matched pair goal. A Matched-Pair is two furniture items of the same
color.  The inference of a Matched-Pair is based on the minor premises
shown in \ex{1}:

\eenumsentence{
\item (Intend A B (Put ?agent ?item-1 ?room-1)
\item (Intend A B (Put ?agent ?item-2 ?room-2)
\item (Equal (Color ?item-1) (Color ?item-2))
}

Making this inference is more demanding for resource limited agents,
than the processing needed in  the Standard task. In the standard
task, in order to agree on one step of the plan, the agents must
access at least one belief about a furniture item they have available.
In order to properly evaluate the option represented by that furniture
item they must access the {\sc warrant} for that option. In contrast,
in both Matched Pair tasks, the {\sc warrant} for both beliefs must be
accessed for both furniture items that could contribute to a
Matched-Pair goal. In addition, the premises in 13 must also be
accessed, and furthermore, each version of the Matched-Pair task
requires one additional premise.

 The difference in the two  Matched Pair tasks is
whether the matches are in the same or different rooms.  Matched-Pair
Same Room requires the additional premise that (Equal ?room-1 ?room-2)
while Matched Pair Two Room requires the additional premise that NOT
(Equal ?room-1 ?room-2).  Because premise 13a is inferred and stored
in memory at the time that a proposal is accepted, and because the
agents always complete one room before starting another, the necessary
premise shown in \ex{0}a is more likely to be salient in the
Matched-Pair-Same-Room task.

As discussed in sections \ref{intro-sec} and \ref{iru-sec}, we wish to
provide a test of hypothesis A2, the discourse inference constraint, and
examine how it affects the coordination of inference in collaborative
planning.  Hypotheses A3 and A5 together imply that the complexity
of inference should interact with the agent's ability to stay
coordinated on inferences. Since our measure of inferential complexity
is the number of independent premises required to draw an inference,
both the Matched-Pair-Same-Room task and the Matched-Pair-Two-Room
task increase inferential complexity.

Evaluating the {\sc quality of solution} for the Matched-Pair tasks
reflects the emphasis on coordinating on inferences, since both
Matched-Pair tasks require that {\bf both} agents make the matched
pair inferences in order to score points for matched-pairs. The task
measures how well agents are {\bf coordinated} on the inferred
intentions that follow from the intentions that were explicitly agreed
upon. Only the intentions that contribute to Matched-Pairs are counted
in the final solution, and the utility of these intentions is the sum
of the utilities of the two furniture items, plus the utility of the
Matched-Pair (50 points).

\begin{figure}[htb]
\centerline{\psfig{figure=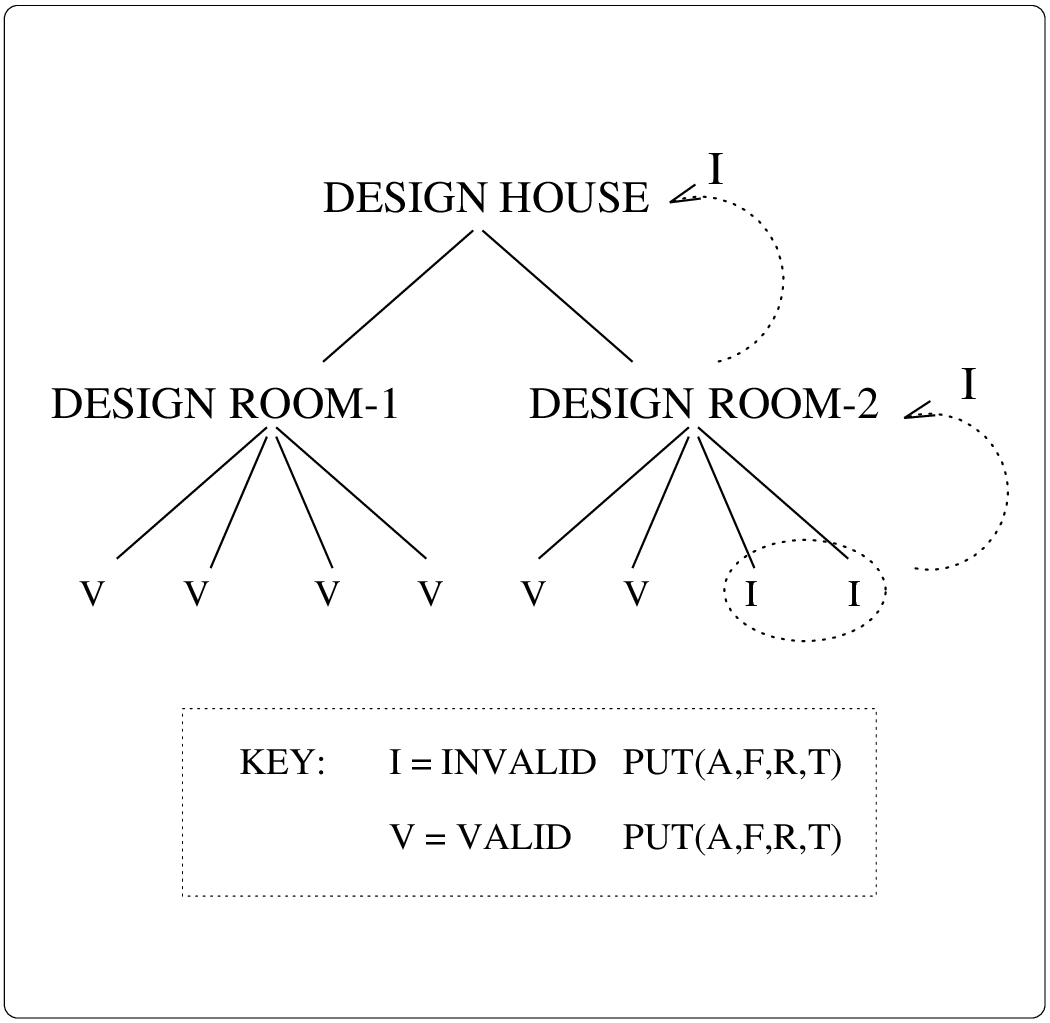,height=2.5in,width=5.0in}}
\caption{In Zero Invalids Task,  invalid steps
invalidate the whole plan. }
\label{task-eval-inval-fig}
\end{figure}

\paragraph{Zero-Invalids Task} The Zero-Invalids task is  fault
intolerant (see figure \ref{task-eval-inval-fig}) The assumption in
the Zero Invalids Task is that any mistake invalidates the whole plan.
This is a feature of task determinacy: while there are still many
possible 8 step plans, all of the solutions with less than 8 items
that would be counted as valid solutions for the Standard Task are not
valid solutions for the Zero Invalids Task.

This task is an example of one extreme of fault intolerance. In
general, how fault tolerant a task is depends on the
interdependency of different subparts of the problem solution.  For
some tasks, a mistake can invalidate the whole solution, for other
tasks, partial solutions without the invalid step may be adequate.
For example, in a task like furnishing a room it may be desirable to
have both a couch and a chair, but if the agents make a mistake and
assume they can use a chair that will end up in a different room, the
room is still partially furnished and usable. On the other hand, in a
task such as building a tower, each step depends on the successful
execution of the previous step and the whole plan may be invalid if a
step to put down a foundation block cannot be executed.

Note that an agent can reject another agent's proposal based on
believing that it would add an invalid step to the plan, as shown in
the rejection utterance act schema  in section \ref{dial-sec}.
Since agents have to agree on each step of the plan, an invalid step
can only be inserted into the plan if both agents have failed to
remember that the preconditions for the plan do not hold.

\subsection{Varying Communicative Strategies}
\label{comm-choice-sec}

Section \ref{model-sec} discussed the discourse act schema that
controls how agents participate in dialogue, and discussed the types
of utterance acts that the discourse acts are composed of. Which
utterance acts a discourse act decomposes into depends on {\sc
communicative strategies} which codify different communicative choices
for how to do a particular discourse action.  Agents are parameterized
for different communicative strategies by placing different expansions
of discourse plans in their plan libraries.

Varying an agent's communicative strategies provides the basis for
testing the hypotheses about the potential benefits of IRUs.  Varying
the degree of explicitness of a discourse act is the basis of the four
communicative strategies tested below: (1) All-Implicit; (2)
Close-Consequence; (3) Explicit-Warrant; and (4)
Matched-Pair-Inference-Explicit.  All of these strategies are
hypothesized to mitigate agents' attentional and inferential resource
limits, under the assumptions about their architecture and the
definition of quality of solution for the task.

Figures \ref{allimp-planop-fig}, \ref{cc-planop-fig},
\ref{expw-planop-fig} and \ref{mpie-planop-fig} show a plan operator
for each strategy. These operators draw on work by Walker and Rambow
\cite{WalkerRambow94}, and also make use of Moser and Moore's
definitions of discourse acts and the integration of Rhetorical
Structure Theory (RST) and Grosz and Sidner's theory of discourse
\cite{MannThompson87,GS86,MoserMoore93,MoserMoore95}.
The predicates in the plan operators are precisely defined by the
collaborative planning model and agent architecture discussed in
section \ref{model-sec}. Each discourse act such as a proposal, is
composed of a {\sc core} act which represents the primary purpose of
the act such as a propose utterance act, as well as a {\sc
contributor} act, such as a warrant, whose purpose is to increase the
likelihood of achieving the intention of the core
\cite{MooreParis93,YMP94,YoungMoore94}.

\paragraph{All-Implicit Strategy}

\begin{figure*}[htb]
\begin{center}
\rule{12cm}{.2mm}
\begin{tabular}{ll}
{\sc name:}  &  Proposal-All-Implicit (?speaker, ?hearer, ?act) \\
{\sc effect:}&   (desire ?hearer (do ?hearer ?act) ?utility-act)\\
{\sc constraints:}     &   (and   (option ?act) \\
     &       (salient ?hearer (utility ?act ?utility-act))) \\
{\sc core:}    &  (propose ?speaker ?hearer ?act)
\end{tabular}

\rule{12cm}{.2mm}

\end{center}
\caption{The {\sc Proposal} plan operator for an All-Implicit Agent}
\label{allimp-planop-fig}
\end{figure*}

The All-Implicit strategy is an expansion of a discourse plan to make
a {\sc proposal}, in which a {\sc proposal} decomposes trivially to
the communicative act of {\sc propose}.  See the plan-operator in
figure \ref{allimp-planop-fig}.  This strategy is the communicative
choice shown in \ref{least-exp-examp} in section \ref{intro-sec}, and
provides a baseline strategy that is consistent with the {\sc
redundancy constraint}. The experiments below will compare the
performance of agents using the All-Implicit strategy with the
performance of agents using the other proposal strategies discussed
below.

In dialogue \ref{bill-kim-dial-examp} on page
\pageref{bill-kim-dial-examp}, both Design-World agents communicate
using the All-Implicit strategy, and the proposals are shown in
utterances 1, 2, and 3. As figure \ref{allimp-planop-fig} shows, the
All-Implicit strategy includes no additional information in proposals,
leaving it up to the other agent to retrieve them from memory.

 The
constraints on using the All-Implicit strategy are that (1) the
proposed ?act is an {\sc option} generated by means-end reasoning and
(2) that the utility is {\sc salient} to the hearer. In the
experiments below, agents are parameterized to use this strategy
consistently, so that an agent using the All-Implicit strategy assumes
everything the hearer knows is always salient. The effect of the
proposal is that the hearer will evaluate that proposal and deliberate
the degree to which the hearer desires the act. However, whether the
hearer will accept or reject the proposal depends on other options the
hearer knows about. Clearly the speaker cannot predict these other
options. Thus the effect of the proposal does not specify that the
action will be intended by the hearer. This holds for all proposal
operators.

The All-Implicit strategy can be used by agents in any of the
Design-World tasks discussed in section \ref{task-def-sec}, since the
agents are capable of making inferences or accessing memory to fill in
what has been left implicit with this strategy.  Other inferences
drawn by the hearer from the proposal utterance act are not shown in
figure
\ref{allimp-planop-fig}.  For example, an agent can use the
All-Implicit strategy in either of the Matched-Pair tasks, leaving it
up to the other agent to infer which other intention makes a match
with the option currently under consideration.

\paragraph{Close Consequence}

\begin{figure*}[htb]
\begin{center}
\rule{12cm}{.2mm}
\begin{tabular}{ll}
{\sc name:}  &  Close-Consequence (?speaker,?hearer,?act)\\
{\sc effect:}&  (and (salient ?hearer (effect ?act ?effect))\\
       &       (bel ?hearer (closed-segment ?act))) \\
{\sc constraints:}&  (and (intend  ?speaker ?hearer ?act) \\
            &  (open-segment ?act))\\
{\sc contributor:}   &  (say ?speaker ?hearer (effect ?act ?effect)) \\
{\sc core:}    &  (close ?speaker ?hearer ?act))
\end{tabular}

\rule{12cm}{.2mm}

\end{center}
\caption{The {\sc Closing} plan operator for a  Close-Consequence Agent}
\label{cc-planop-fig}
\end{figure*}

In dialogue \ex{1}, agent CLC uses the Close-Consequence strategy.  The
plan-operator for this strategy is shown in figure
\ref{cc-planop-fig}.  The core of the strategy is
explicit {\sc closing} statements, such as \ex{1}-2, on the
completion of the intention associated with a discourse segment.
A contributor to
CLC's {\sc closing} discourse act is an IRU such as \ex{1}-3: CLC
makes the inference explicit that since they have
agreed on putting the green rug in the study, they no longer have the
green rug (act-effect inference).

\begin{small}
\eenumsentence
{
\item[1:]
BILL: {\it Then, let's put the green rug in the study.} \\
(propose agent-bill agent-clc option-30: put-act (agent-bill green
rug room-1))

\item[2:]
CLC: {\it So, we've agreed to put the green rug in the study.} \\
(close agent-clc agent-bill intended-30: put-act (agent-bill green
rug room-1))

\item[3:]
CLC: {\sc And we no longer have green rug.} \\
(say agent-clc agent-bill bel-48: hasn't (agent-bill green rug))

\label{ccl-dial}
}
\end{small}

The Close-Consequence strategy of making inferences explicit at the
close of a segment models the naturally occurring
example in \ref{elig-examp}.  In both cases an inference is made
explicit that follows from what has just been said, and the inference
is sequentially located at the close of a discourse segment.
This strategy can be used by agents in any of the
Design-World tasks discussed in section \ref{task-def-sec}.

The Close-Consequence strategy will be used to test hypothesis C2
about potential benefits of making inferences explicit, and will be
contrasted with the All-Implicit strategy where no closing acts are
produced. Note in dialogue \ref{bill-kim-dial-examp} on page
\pageref{bill-kim-dial-examp} that both the agents go on to the next
phase of the plan, leaving the inference of both Acceptance and
Closing for the other agent to make.  However, Close-Consequence is
not a good test of other hypotheses because in the experiments both
agents always make act-effect inferences, and these inferences are not
difficult to make. See \cite{Walker95} for a discussion of experiments
which vary an agent's capability to make these inferences.

\paragraph{Explicit Warrant}

\begin{figure*}[htb]
\begin{center}
\rule{12cm}{.2mm}
\begin{tabular}{ll}
{\sc name:}  &  Proposal-Explicit-Warrant (?speaker, ?hearer, ?act)\\
{\sc effect:}& (and (desire ?hearer (do ?hearer ?act) ?utility-act)\\
       &       (salient ?hearer (utility ?act ?utility-act))) \\
{\sc constraints:}&  (and  (option ?act) \\
     &       (not-salient ?hearer (utility ?act ?utility-act))) \\
{\sc contributor:}   &   (say ?speaker ?hearer (utility ?act ?utility-act)) \\
{\sc core:}    &  (propose ?speaker ?hearer ?act)
\end{tabular}

\rule{12cm}{.2mm}

\end{center}
\caption{The {\sc Proposal} plan operator for an Explicit-Warrant Agent}
\label{expw-planop-fig}
\end{figure*}

The Explicit-Warrant strategy varies the proposal discourse act by
including {\sc warrant} IRUs in each proposal. The plan
operator is given in figure
\ref{expw-planop-fig} and exemplified by the dialogue excerpt
in \ex{1}. Remember that a {\sc warrant} for an
intention is a reason for adopting the intention, and here {\sc
warrants} are the score propositions that give the utility of the
proposal, which are mutually believed at the outset of the dialogues.
In \ex{1}, the {\sc warrant} IRU in \ex{1}-1 contributes
to the proposal (core act) in \ex{1}-2.


\begin{small}
\eenumsentence
{
\item[1:]
IEI: {\sc  Putting in the green rug is worth  56} \\
(say agent-iei agent-iei2 bel-2: score  (option-2:  put-act
(agent-iei green rug room-1) 56))

\item[2:]
IEI: {\it Then, let's put the green rug in the study.} \\
(propose agent-iei agent-iei2 option-2:  put-act (agent-iei green rug
room-1))
}
\end{small}

The plan operator in figure \ref{expw-planop-fig} specifies that an
effect of using this plan is that the utility of the proposal option
is salient. Since warrants are used by the other agent in
deliberation, the Explicit-Warrant strategy can save the other agent
the processing involved with determining which facts are relevant for
deliberation and retrieving them from memory. A constraint on using the
Explicit-Warrant plan operator is that the utility of the proposal act
is not already salient.

In the experiments below, agents are parameterized to use this strategy
consistently, with the result that an agent using the Explicit-Warrant
strategy assumes that the warrant is never salient for the hearer.
See \cite{JordanWalker95} for experiments in which an agent attempts to
maintain a dynamic model of what is salient for the other agent. The
Explicit-Warrant strategy also occurs in natural dialogues as shown in
the naturally occurring example in dialogue
\ref{walnut-examp}.

This strategy can be used by agents in any of the Design-World tasks
discussed in section \ref{task-def-sec}. The Explicit Warrant strategy
provides a test of hypothesis A1: agents produce Attention IRUs to support
the processes of deliberating beliefs and intentions. It can also be
used to test hypothesis A4: the choice to produce an Attention IRU is
related to the degree to which an agent is resource limited in
attentional capacity.  In the Standard task this is predicted to
improve the performance of resource limited agents. In the
Zero-NonMatching beliefs task this strategy should increase the
likelihood that agents coordinate their beliefs about the warrants
underlying different plan steps.

\paragraph{The Matched-Pair-Inference-Explicit strategy}

\begin{figure*}[htb]
\begin{center}
\rule{12cm}{.2mm}
\begin{tabular}{ll}
{\sc name:}  &  Proposal-Matched-Pair (?speaker, ?hearer, ?act1)\\
{\sc effect:}& (and (desire ?hearer (do ?hearer ?act1) ?utility-act)\\
       & (salient ?hearer (intend ?speaker ?hearer ?act2))) \\
{\sc constraints:}&  (and (option ?act1) \\
            &       (matched-pair ?act1 ?act2) \\
     &       (salient ?hearer (utility ?act1 ?utility-act))) \\
   &       (not (salient ?hearer (intend ?speaker ?hearer ?act2))) \\

{\sc contributor:}   & (say ?speaker ?hearer (intend ?speaker ?hearer ?act2))
\\
{\sc core:}    &  (propose ?speaker ?hearer ?act1)
\end{tabular}

\rule{12cm}{.2mm}

\end{center}
\caption{The {\sc Proposal} plan operator for an
Matched-Pair-Inference-Explicit Agent}
\label{mpie-planop-fig}
\end{figure*}

The Matched-Pair-Inference-Explicit strategy expands the {\sc
proposal} discourse act to two communicative acts. See figure
\ref{mpie-planop-fig}. The contributor of the proposal consists of
statement about what is already intended, while the core is a propose
utterance act, as in \ex{1}-6 followed by
\ex{1}-7 in one turn:\footnote{The names of agents who use the
Matched-Pair-Inference-Explicit strategy are a numbered version of the
string ``IMI'' which stands for Implicit acceptance, Match Inference.}

\begin{small}
\eenumsentence{
\item[6:]
IMI2: {\sc We agreed to put the purple couch in the study.} \\
(say agent-imi2 agent-imi intended-51: put-act (agent-imi2 purple couch
room-1))

\item[7:]
IMI2: {\it Then, let's put the purple rug in the living room.} \\
(propose agent-imi2 agent-imi option-80: put-act (agent-imi2 purple rug
room-2))

}
\end{small}

The statement in \ex{0}-6 is an IRU, because it realizes information
previously inferred by both agents, and models the IRU in dialogue
\ref{certif-examp}.  Matched-Pair-Inference-Explicit is the variation
discussed in section \ref{intro-sec} in choice \ref{more-exp-examp}.
This strategy is only intended to be used in the Matched-Pair tasks as
a way of testing hypothesis A2, the discourse inference constraint.
As figure \ref{mpie-planop-fig} shows, a constraint on using this plan
is that the speaking agent has inferred a Matched-Pair for the option
being proposed.

Although this strategy is specifically tied to Matched-Pair
inferences, it  provides a test of a general strategy for
making premises for inferences salient, in tasks that
are inferentially complex, and which also require agents to remain
coordinated on inferences. For example, to generalize this strategy to
other cases of plan-related inferences, the clauses for (Matched-Pair
?act1 ?act2) could be replaced with the more general (Generates ?act1
$\wedge$ ?act2 ?act3), where the generates relation is to be inferred
\cite{Pollack86a,Dieugenio93}.

Note that the effect of using this strategy is not that the hearer
makes the matched pair inference, rather the effect is that the
premise for the desired inference is salient. A constraint on using
this strategy is that this premise is not already salient. However,
agents parameterized with this strategy always assume that the premise
is not salient for the hearer. See \cite{JordanWalker95} for experiments
in which agents attempt to maintain a dynamic model of the other
agent's attentional state.

\subsection{Plan Evaluation}
\label{dw-plan-eval-sec}

\begin{figure}[htb]
\centerline{\psfig{figure=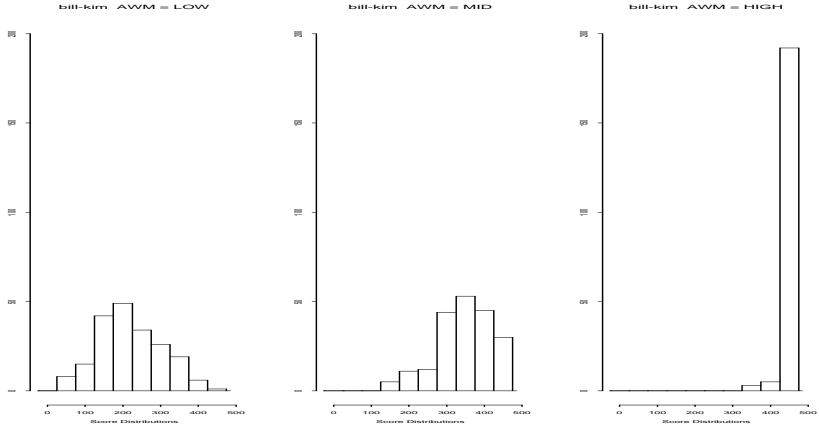,height=2.25in,width=4.5in}}
\caption{Performance Distributions showing
the effect of {\sc awm} parameterization for dialogues between
two All-Implicit Agents  when all processing is free. The three
performance
distributions are for {\sc low, mid} and {\sc high awm} agents.}
\label{bill-kim-hist-fig}
\label{baseline-fig} \end{figure}

\begin{figure}[htb]
\centerline{\psfig{figure=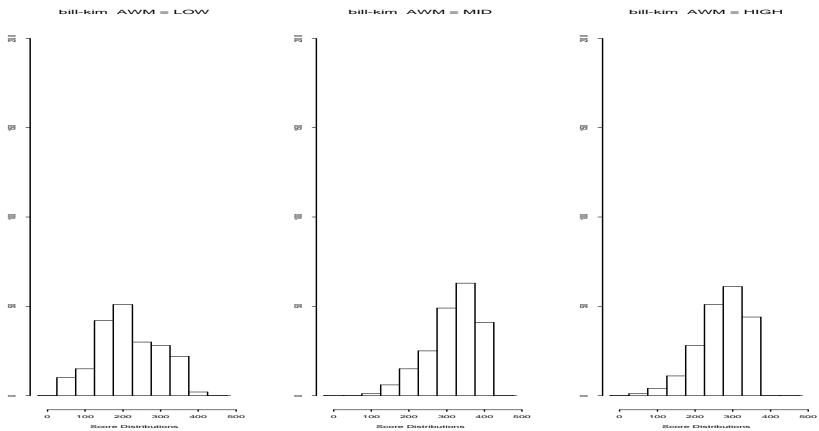,height=2.25in,width=4.5in}}
\caption{Performance distributions showing the effect of increased
retrieval cost
for each {\sc awm} range for dialogues between two All-Implicit Agents. The
three
performance
distributions are for {\sc low, mid} and {\sc high awm} agents. retcost = .001}
\label{retcost-baseline-fig} \end{figure}

Section \ref{model-plan-eval-sec} specified a model of how
collaborative plans are evaluated in terms of {\sc quality of
solution} and {\sc collaborative effort}. Design-World is constructed
in order to be able to measure the quality of a solution as well as
collaborative effort. Section \ref{task-def-sec} defined quality of
solution for all of the Design-World tasks.
We want to  examine
trade-offs in performance between strategy choices.

It is obvious that these trade-offs can be related to the relative
contributions of total cost of communication versus the total cost of
inference versus the total cost of retrieval for both agents'
collaborative effort.  Thus, to calculate collaborative effort, we
cannot simply add up the number of retrievals, inferences and
messages.  Consider that a Consequence IRU that makes an inference
explicit ensures that an inferred belief
becomes part of the discourse model. However, if the inference would
have been made anyway, the benefit of the strategy is dependent upon
whether the effort to make the inference without the consequence IRU
would have been greater than the cost of processing the extra
utterance of the Consequence IRU.  A similar argument holds for the
potential benefit of Attention IRUs.  Whenever an Attention IRU
reduces overall effort for retrieval while not increasing
communication effort to the same degree, it will be beneficial. This
hypothesis is given below in a general form.

\begin{quote}
HYPOTH-I1: Strategies that reduce collaborative effort without
affecting quality of solution
are beneficial.
\end{quote}

This hypothesis follows directly from the definition
of performance repeated here for convenience from section
\ref{model-plan-eval-sec}:

\begin{quote}
{\sc performance} $=$
  {\sc quality of solution} -- {\sc collaborative effort}
\end{quote}

We need to introduce parameters for the effort involved with each of
the component processes because they are not strictly comparable, and
because these modules are implementation dependent.  Thus agents'
retrieval, inference and communicative costs are parameterized by (1)
{\sc commcost}: cost of sending a message; (2) {\sc infcost}: cost of
inference; and (3) {\sc retcost}: cost of retrieval from memory.
Collaborative effort is then defined as:

\begin{quote}
\begin{tabular}{l}
{\sc collaborative effort} $=$  \\
({\sc commcost} $\times$ total messages for both agents)\\
 + ({\sc infcost} $\times$ total inferences for both agents)\\
 + ({\sc retcost} $\times$ total retrievals for both agents)
\end{tabular}
\end{quote}

We will use these cost parameters to explore three extremes in this
space: (1) when processing is free; (2) when retrieval effort
dominates other processing costs; and (3) when communication effort
dominates other processing costs.  The parameters support modeling
various instantiations of the agent architecture given in figure
\ref{irma-fig}.  For example, varying the cost of retrieval models
different assumptions about how the beliefs database, plan library and
working memory are implemented.  Varying the cost of communication
models situations in which communication planning is very costly. The
relation between the values of these parameters and the utilities of
the steps in the plan determines experimental outcomes, rather than
the absolute values.

As an example of the effect of varying these costs, consider the
plots of performance distributions shown in figures \ref{baseline-fig} and
\ref{retcost-baseline-fig} for {\sc low, mid} and {\sc high awm}.
In these figures, performance is plotted on the x-axis and number of
simulations at that performance level are given by bars on the y-axis.
The performance distributions in figure
\ref{baseline-fig} demonstrate the increase in {\sc quality of
solution} that we would expect with increases in {\sc awm}, given no
processing costs.\footnote{These distributions approximate Beta
distributions
\cite{Wilks62}, and this approximation was used to determine that 200
runs would guarantee stable results.  The Beta distribution with the
largest variance, for parameters R and S greater than or equal to 1,
is the uniform distribution. This largest variance distribution would
require approximately 133 samples \cite{Siegel56,Wilks62}.  An
empirical evaluation of the adequacy of this sample size for three
different strategies was tested to see if any differences showed up in
alternate runs of 100; no differences were found.} Figure
\ref{retcost-baseline-fig} shows what happens when processing is not
free: here a retrieval cost of .001 means that every memory access
reduces quality of solution by 1/1000 of a point (remember that the
utilities of plan steps range between 10 and 56).  As figure
\ref{retcost-baseline-fig} shows, the ability to access the whole
beliefs database in reasoning does not always improve performance
since {\sc high awm} agents perform similarly to {\sc mid awm} agents.

\subsection{Summary: Mapping from Naturally Occurring Data to Design World
Experiments}

Section \ref{iru-sec} proposed hypotheses about the function of IRUs
in human to human collaborative planning dialogues, and then section
\ref{model-sec} presented a model for collaborative planning dialogues
based on the observations in section \ref{iru-sec}.  Section
\ref{dw-sec} then described Design-World as a testbed of the model,
and sections \ref{task-def-sec} and \ref{comm-choice-sec} introduced a
number of parameters of the testbed that are intended to model the
features of the human-human dialogues and support testing of the
hypotheses.  Here I wish to summarize the mapping between the
naturally occurring dialogues and the design of the testbed in order
to clarify the basis for the experiments in the next section.

The testbed and the experimental parameters are based on the following
mapping between human-human collaborative planning dialogues and the
testbed. First, the planning and deliberation aspects of human
processing are modeled with the IRMA architecture, and resource limits
on these processes are modeled by extending the IRMA architecture with
a model of Attention/Working Memory ({\sc awm}) which has been shown to
model a limited but critical set of properties of human processing.
Second, the processing of dialogue is tied to the agent architecture.
Third, the mapping of a {\sc warrant} relation between an act and a
belief in naturally occurring examples such as \ref{walnut-examp} is
modeled with a {\sc warrant} relation between an act and a belief in
Design-World as seen in the Explicit-Warrant communication strategy in
section \ref{comm-choice-sec}.  Fourth, the mapping assumes that
arbitrary content based inferences in natural dialogues such as that
discussed in relation to example \ref{certif-examp} can be mapped to
content based inferences in Design-World such as those required for
doing well on the Matched-Pair tasks. Fifth, the mapping is based on
the assumption that task difficulty in naturally occurring tasks such
as those in the financial advice domain can be related to three
abstract features: (1) inferential complexity as measured by the
number of premise required for making an inferences; (2) degree of
belief coordination required on intentions, inferences and beliefs
underlying a plan; and (3) task determinacy and fault tolerance.
Finally the mapping assumes that it is reasonable to evaluate the
performance of the agents in collaborative planning dialogues by using
domain plan utility for a measure of the quality of solution and
defining the cost to achieve that solution as collaborative effort,
appropriately parameterized.

The details of this mapping specifies how the testbed implements the
model of collaborative planning and provides the basis for
extrapolating from the testbed experimental results to the human-human
dialogues that are being modeled. The testbed provides an excellent
environment for testing the hypotheses to the extent that the model
captures critical aspects of human-human dialogues.

\section{Experimental Results}
\label{results-sec}

\subsection{Statistically Evaluating  Performance}

The experiments examine the interaction between tasks, communication
strategies and {\sc awm} resource limits.  Every experiment varies {\sc awm}
over
three ranges: {\sc low, mid}, and {\sc high}.  In order to run an
experiment on a particular communicative strategy for a particular
task, 200 dialogues for each {\sc awm} range are simulated.  Because the {\sc
awm}
model is probabilistic, each dialogue simulation has a different
result.  The {\sc awm} parameter yields a performance distribution for very
resource limited agents ({\sc low}), agents hypothesized to be similar
to human agents ({\sc mid}), and resource unlimited agents ({\sc
high}).  Sample performance distributions for {\sc quality of
solution} (with no collaborative effort subtracted) from runs of two
All-Implicit agents for each {\sc awm} setting are shown in figure
\ref{bill-kim-hist-fig}.

To test our hypotheses, we want to {\bf compare} the performance of
two different communicative strategies for a particular task, under
different asssumptions about resource limits and processing costs.  To
see the effect of communicative strategy and {\sc awm} over the whole range
of {\sc awm} settings, we first run a two-way analysis of variance (anova)
with {\sc awm} as one factor and communication strategy as
another.\footnote{The experimental performance distributions are not
normal and the variance is not the same over different samples,
however anova is robust against the violation of these assumptions
under the conditions in these experiments
\cite{Cohen95,Keppel73}.} The anova tells us whether: (1) {\sc awm} alone
is a significant factor in predicting performance; (2) communication
strategy alone is a significant factor in predicting performance; and
(3) whether there is an interaction between communication strategy and
{\sc awm}.

However, anova alone does not enable us to determine the particular {\sc awm}
range at which a communication strategy aids or hinders performance,
and many of the hypotheses about the benefits of particular
communication strategies are specific to how resource limited an agent
is. Furthermore, whenever strategy affects performance positively for
one value of {\sc awm} and negatively for another value of {\sc awm}, the
potential effects of strategy cannot be seen from the anova alone.
Therefore, we conduct planned comparisons of strategies using the
modified Bonferroni test (hereafter MB) \cite{Keppel73}, within each
{\sc awm} range setting to determine which {\sc awm} range the strategy affects
\cite{Cohen95,Keppel73}.\footnote{According to the modified Bonferroni
test, the significant F values for the planned comparisons reported
below are 3.88 for a p $<$ .05, 5.06 for a p $<$ .025, 6.66 for a p $<$
.01, and 9.61 for a p $<$ .002.} On the basis of these comparisons we
can say whether a strategy is {\sc beneficial} for a particular task
for a particular {\sc awm} range.

\begin{quote}

A strategy A is {\sc beneficial} as compared to a strategy B, for a
particular {\sc awm} range, in the same task situation, with the same cost
settings, if the mean of A is significantly greater than the mean of
B, according to the modified Bonferroni test (MB) test.
\end{quote}

The converse of {\sc beneficial} is {\sc detrimental}:

\begin{quote} A strategy A is {\sc detrimental} as compared to a
strategy B, for a particular {\sc awm} range, in the same task situation,
with the same cost settings, if the mean of A is significantly less
than the mean of B, according to the modified Bonferroni test (MB)
test.  \end{quote}

Strategies need not be either {\sc beneficial} or {\sc detrimental},
there may be no difference between two strategies. Also with the
definition given above a strategy may be both {\sc beneficial} and
{\sc detrimental} depending on the range of {\sc awm} that the two
strategies are compared over, i.e.  A strategy may be beneficial for
{\sc low awm} agents and detrimental for {\sc high awm} agents.

A {\sc difference plot} such as that in figure
\ref{free-ret-iei-fig} is used to summarize a comparison of two
strategies, strategy 1 and strategy 2.  In the comparisons below,
strategy 1 is either Close-Consequence,\footnote{In experiments with
Close-Consequence only one agent in a dialogue uses the
Close-Consequence strategy because the use of this strategy is
constrained to when the dialogue segment is open. See figure
\ref{cc-planop-fig}. Since only one agent will ever produce a closing
statement for any dialogue segment, only one agent is given the option
in the simulations.} Explicit-Warrant, or
Matched-Pair-Inference-Explicit and strategy 2 is the All-Implicit
strategy.  {\bf Differences} in performance means between two
strategies are plotted on the Y-axis against {\sc awm} ranges on the
X-axis.  Each point in the plot represents the difference in the means
of 200 runs of each strategy at a particular {\sc awm} range.  These
plots summarize the information from 1200 simulated dialogues.

\subsection{Standard Task}

\begin{figure}[htb]
\centerline{\psfig{figure=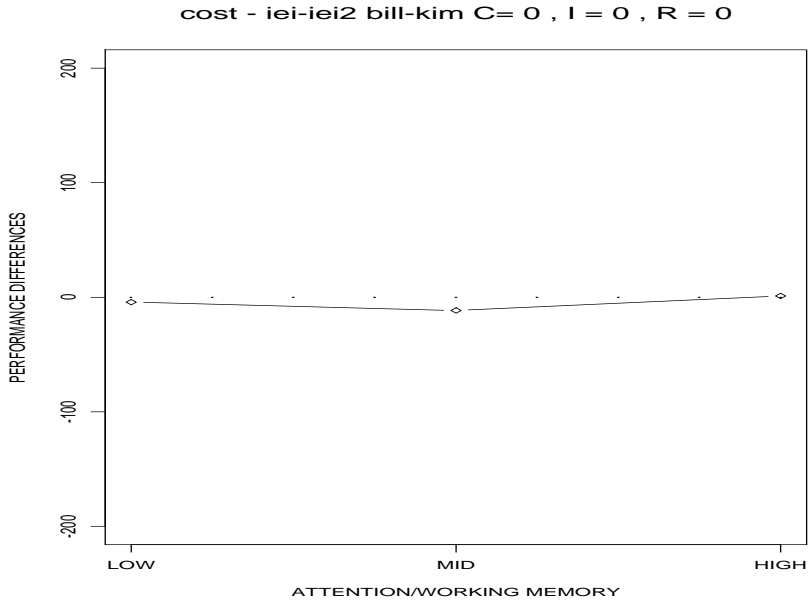,height=3.25in,width=4.5in}}
\caption{If Processing is Free, Explicit-Warrant is neither beneficial
nor detrimental for all {\sc awm} settings: Strategy 1 of two Explicit-Warrant
agents and
strategy 2 of two All-Implicit agents: Task = Standard, commcost = 0,
infcost = 0, retcost = 0}
\label{free-ret-iei-fig}
\end{figure}

\begin{figure}[htb]
\centerline{\psfig{figure=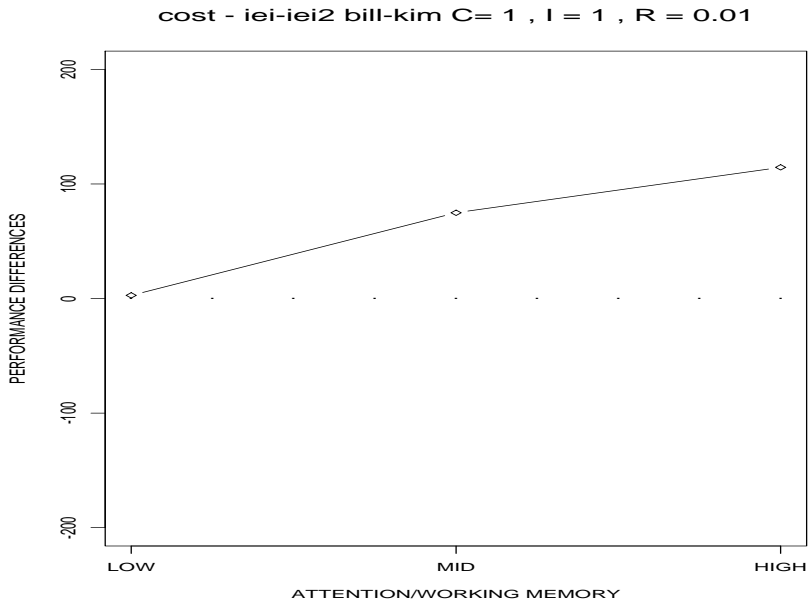,height=3.25in,width=4.5in}}
\caption{Explicit-Warrant is beneficial for  {\sc mid} and
{\sc high awm} agents when  Retrieval dominates processing costs: Strategy 1 is
two Explicit-Warrant agents and strategy 2 is two All-Implicit agents:
Task = Standard, commcost = 1, infcost = 1, retcost = .01}
\label{ret-iei-fig}
\end{figure}

\begin{figure}[htb]
\centerline{\psfig{figure=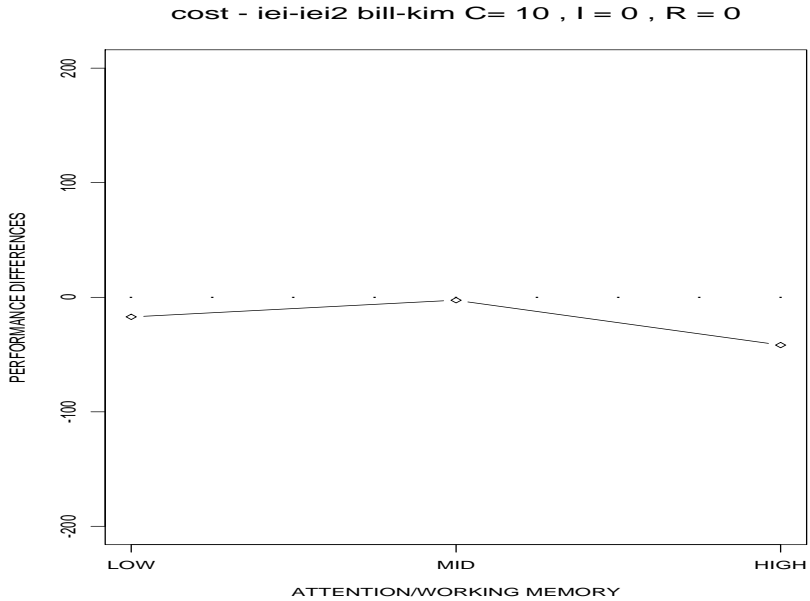,height=3.25in,width=4.5in}}
\caption{Explicit-Warrant is detrimental for {\sc low} and {\sc high awm}
agents
when communication effort is high:
Strategy 1 is two Explicit-Warrant agents and strategy 2 is two
All-Implicit agents: Task = Standard, commcost = 10, infcost = 0,
retcost = 0}
\label{exp-iei-fig}
\end{figure}

Remember that the Standard task is defined so that the {\sc quality of
solution} that agents achieve for a {\sc design-house} plan,
constructed via the dialogue, is the sum of the utilities of each
valid step in their plan. The task has multiple correct solutions and
is fault tolerant because the point values for invalid steps in the
plan are simply subtracted from the score, with the effect that agents
are not heavily penalized for making mistakes. Furthermore, the task
has low inferential complexity: the only inferences agents are
required to make are those for deliberation and means-end reasoning.
In both of these cases, to make these inferences, agents are only
required to access a single minor premise

All-Implicit agents do fairly well at the Standard task, under
assumptions that all processing is free, as shown in the performance
plot in figure \ref{baseline-fig}. However, as retrieval costs
increase, {\sc high awm} agents don't do as well as when retrieval is
free, because they expend too much effort on retrieval during
collaborative planning.  Compare the {\sc high awm} distribution in
figure \ref{baseline-fig} with that in figure
\ref{retcost-baseline-fig}.  Thus for the Standard task, {\sc high
awm} agents have the potential to benefit from communication
strategies that reduce the total effort for retrieval, when retrieval
is not free. In addition, although the task has minimal inferential
complexity, easy access to information that is used for deliberation,
which the Explicit-Warrant strategy provides,
could benefit {\sc low awm} agents, since they might otherwise make
nonoptimal decisions. Furthermore, although the task is fault
tolerant, agents are still penalized for making errors since errors do
not contribute to performance. Thus for the Standard task,
communication strategies such as Close-Consequence that can reduce the
number of errors could be beneficial. Below we will compare the
All-Implicit strategy to the Explicit-Warrant strategy and the
Close-Consequence strategy.

\paragraph{Explicit-Warrant}
The Explicit-Warrant strategy can be used in the Standard task to test
hypothesis A1: agents produce Attention IRUs to support the processes
of deliberating beliefs and intentions. It can also be used to test
hypothesis A4: the choice to produce an Attention IRU is related to
the degree to which an agent is resource limited in attentional
capacity.  Thus one prediction is that the Explicit-Warrant strategy
will result in higher performance for {\sc low awm} agents even when
processing is free by ensuring that they can access the warrant and
use it in deliberation, thus making better decisions.

Figure \ref{free-ret-iei-fig} plots the differences in the performance
means between the Explicit-Warrant strategy and the All-Implicit
strategy for {\sc low, mid} and {\sc high awm} agents.  A two-way
anova exploring the effect of {\sc awm} and the Explicit-Warrant
strategy for the Standard task shows that {\sc awm} has a large effect
on performance (F= 336.63, p$<$ .000001). There is no main effect for
communicative strategy (F = 1.92, p $<$ 0.16). However, there is an
interaction between {\sc awm} and communicative choice (F=1136.34,
p$<$ .000001).

By comparing performance within a particular {\sc awm} range for each
strategy we can see which {\sc awm} settings interact with
communicative strategy.  The planned comparisons using the Modified
Bonferonni ({\sc mb}) test show that the Explicit-Warrant strategy is
neither beneficial nor detrimental in the Standard task, in comparison
with the All-Implicit strategy, if all processing is free ({\sc
mb(low)} = 0.29, ns; {\sc mb(mid)} = 2.79, ns; {\sc mb(high)} = 0.39,
ns).  Note that there is a trend towards the Explicit-Warrant strategy
being detrimental at {\sc mid awm}.

 The hypothesis based on the corpus analysis was that {\sc low awm}
agents might benefit from communicative strategies that include IRUs.
However, this hypothesis is disconfirmed. Further analysis of this
result suggests a hypothesis not apparent from the corpus analysis:
any beneficial effect of an IRU can be cancelled for resource limited
agents because IRUs may displace other information from working memory
that is more useful.  In this case, despite the fact that the warrant
information is useful for deliberation, making the warrant salient
displaces information that can be used to generate other options.
When agents are very resource-limited making an optimal decision is
not as important as being able to generate multiple options.

The Explicit-Warrant strategy can also be used in the Standard task to
test hypothesis I1: strategies that reduce collaborative effort
overall may be beneficial. Thus, another prediction is that by
providing the warrant used in deliberating a proposal with every
proposal, the Explicit-Warrant strategy has the potential to reduce
resource consumption when accessing memory has some processing cost.

Figure \ref{ret-iei-fig} plots the differences in the performance
means between the Explicit-Warrant strategy and the All-Implicit
strategy for {\sc low, mid} and {\sc high awm} agents when retrieval
effort dominates processing.  A two-way anova exploring the effect of
{\sc awm} and the Explicit-Warrant strategy for the Standard task, when
retrieval cost dominates processing, shows that {\sc awm} has a large effect
on performance (F= 330.15, p$<$ .000001). There is also a main effect
for communicative strategy (F = 5.74, p $<$ 0.01), and an interaction
between {\sc awm} and communicative choice (F= 1077.64, p$<$ .000001).

The planned comparisons using the MB test to compare performance at
each {\sc awm} range show that, in the Standard task, in comparison with the
All-Implicit strategy, the Explicit-Warrant strategy is neither
beneficial nor detrimental for {\sc low awm} agents ({\sc mb(low)} = 0.27,
ns). However, hypothesis I1 is confirmed because the Explicit-Warrant
strategy is beneficial for {\sc mid awm} agents {\sc mb(mid)} = 86.43, p$<$
.002. The Explicit-Warrant strategy also tends towards improving
performance for {\sc high awm} agents {\sc mb(high)} = 2.07, p $<$ .10).
For higher {\sc awm} values, this trend is because the beliefs
necessary for deliberating the proposal are made available in the
current context with each proposal, so that agents don't have to
search memory for them.

As an additional test of hypothesis I1, a final experiment tests the
Explicit-Warrant strategy against the All-Implicit strategy in a
situation where the cost of communication dominates other processing
costs.  Figure \ref{exp-iei-fig} plots the differences in the
performance means between the Explicit-Warrant strategy and the
All-Implicit strategy for {\sc low, mid} and {\sc high awm} agents
when communication effort dominates processing.  A two-way anova
exploring the effect of {\sc awm} and the Explicit-Warrant strategy for the
Standard task, when communication effort dominates processing, shows
that {\sc awm} has a large effect on performance (F=  409.52, p$<$ .000001).
There is also a main effect for communicative strategy (F = 28.12, p
$<$ 0.000001), and an interaction between {\sc awm} and communicative choice
(F=
960.24, p$<$ .000001).

The planned comparisons using the MB test to compare performance at
each {\sc awm} range show that in this situation, when communication effort
dominates processing, the Explicit-Warrant strategy is neither
beneficial nor detrimental for {\sc mid awm} agents ({\sc mb(mid)} = 0.12,
ns. However, the Explicit-Warrant strategy is detrimental for both
{\sc low} and {\sc high awm} agents, {\sc mb(low)} = 7.69, p$<$ .01;
{\sc mb(high)} = 39.65, p $<$ .01). Since this strategy includes an extra
utterance with every proposal and provides no clear benefits, it is
detrimental to performance in the Standard task when communication
effort dominates processing. Below, when we compare this situation
with that in the Zero-NonMatching-Beliefs task, we will see that this
is due to the fact that the Standard task has low coordination
requirements.

\begin{figure}[htb]
\centerline{\psfig{figure=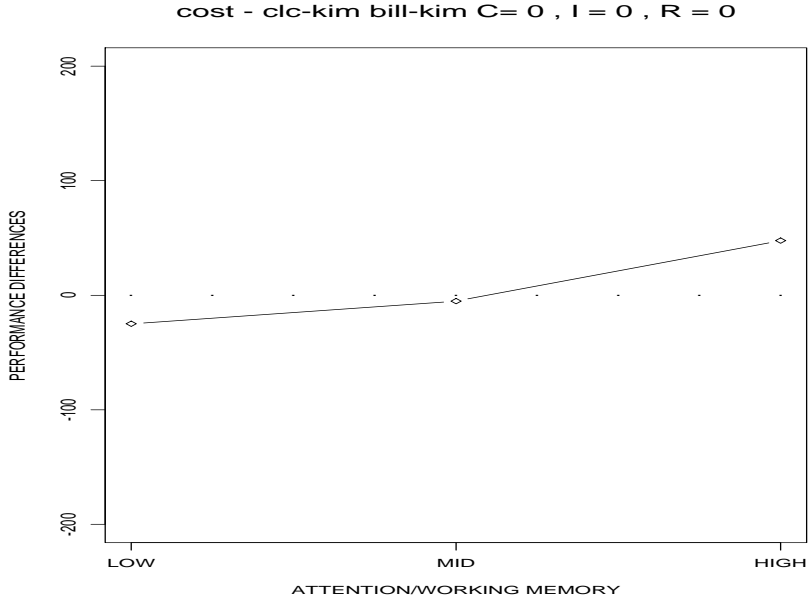,height=3.25in,width=4.5in}}
\caption{Close-Consequence can be detrimental in the Standard Task
for {\sc low awm} agents and beneficial for {\sc high awm} agents.
Strategy 1 is the combination of an All-Implicit agent with a
Close-Consequence agent and Strategy 2 is two All-Implicit agents,
Task = Standard, commcost = 0, infcost = 0, retcost = 0}
\label{cost-clc-kim-fig} \end{figure}

\paragraph{Close-Consequence}

The Close-Consequence strategy of making inferences explicit can be
used in the Standard task to test hypothesis C4: the choice to produce a
Consequence IRU is related to a measure of `how important' the
inference is.  Even though the Standard task is fault tolerant, every
invalid step reduces the quality of solution of the final plan.
Making act-effect inferences explicit decreases the likelihood of
making this kind of error.

The difference plot in figure
\ref{cost-clc-kim-fig} plots performance differences between the
Close-Consequence strategy and the All-Implicit strategy, in the
Standard task, when all processing is free.  A two-way anova exploring
the effect of {\sc awm} and the Close-Consequence strategy in this situation
shows that {\sc awm} has a large effect on performance (F= 249.20, p$<$
.000001), and that there is an interaction between {\sc awm} and
communicative choice (F= 919.27, p$<$ .000001).

Planned comparisons between strategies for each {\sc awm} range shows that
the Close-Consequence strategy is detrimental in comparison with
All-Implicit for {\sc low awm} agents ({\sc mb(low)} = 8.70, p $<$ .01).
This is because generating options contributes more to performance for
agents with {\sc low awm} than avoiding errors, and the additional
utterances that make inferences explicit in the Close-Consequence
strategy has the effect of displacing facts that could be used in
means end reasoning to generate options. There is no difference in
performance for {\sc mid awm} agents ({\sc mb(mid)} = .439, ns).

However, comparisons between the two strategies for {\sc high awm}
agents shows that the Close-Consequence strategy is beneficial in
comparison with All-Implicit ({\sc mb(high)} = 171.71, p $<$ .002).  See
figure \ref{cost-clc-kim-fig}.  This is because the belief
deliberation algorithm increases the probability of {\sc high awm}
agents choosing to believe out of date beliefs about the state of the
world.  The result is that they are more likely to have invalid steps
in their plans. Thus the Close-Consequence strategy is beneficial
because reinforcing the belief that a furniture item has been used
makes it less likely that agents will believe that they still have
that furniture item. This result is not predicted by any hypotheses,
but as discussed in section \ref{irma-sec}, this property of the
belief deliberation mechanism has some intuitive appeal.  In any case,
this result provides a data point for the benefit of a strategy for
making inferences explicit when the probability of making an error
increases if that inference is not made.

\begin{figure}[htb]
\centerline{\psfig{figure=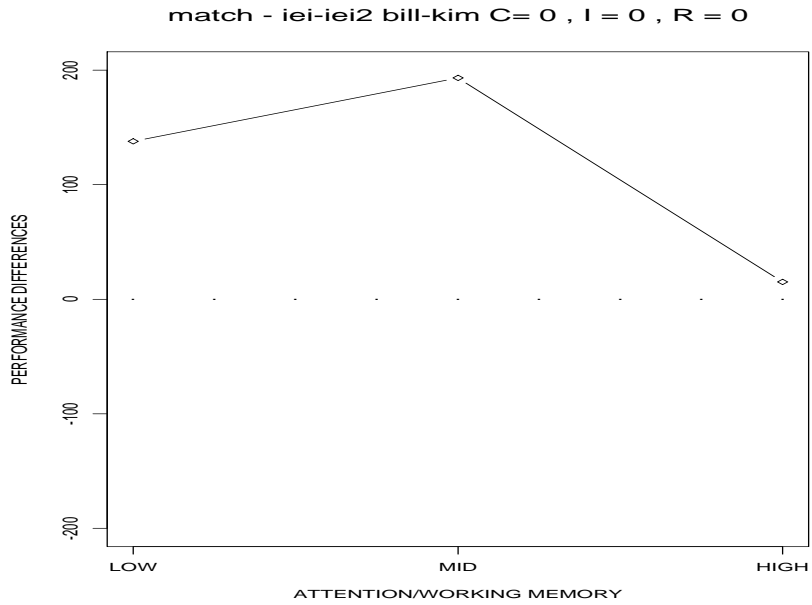,height=3.25in,width=4.5in}}
\caption{Explicit-Warrant is beneficial for Zero-NonMatching-Beliefs
Task for {\sc low} and {\sc mid awm} agents: Strategy 1 is two Explicit-Warrant
agents and strategy 2 is two
All-Implicit agents: Task = Zero-Nonmatching-Beliefs, commcost = 0,
infcost = 0, retcost = 0}
\label{iei-nmb-fig}
\end{figure}

\begin{figure}[htb]
\centerline{\psfig{figure=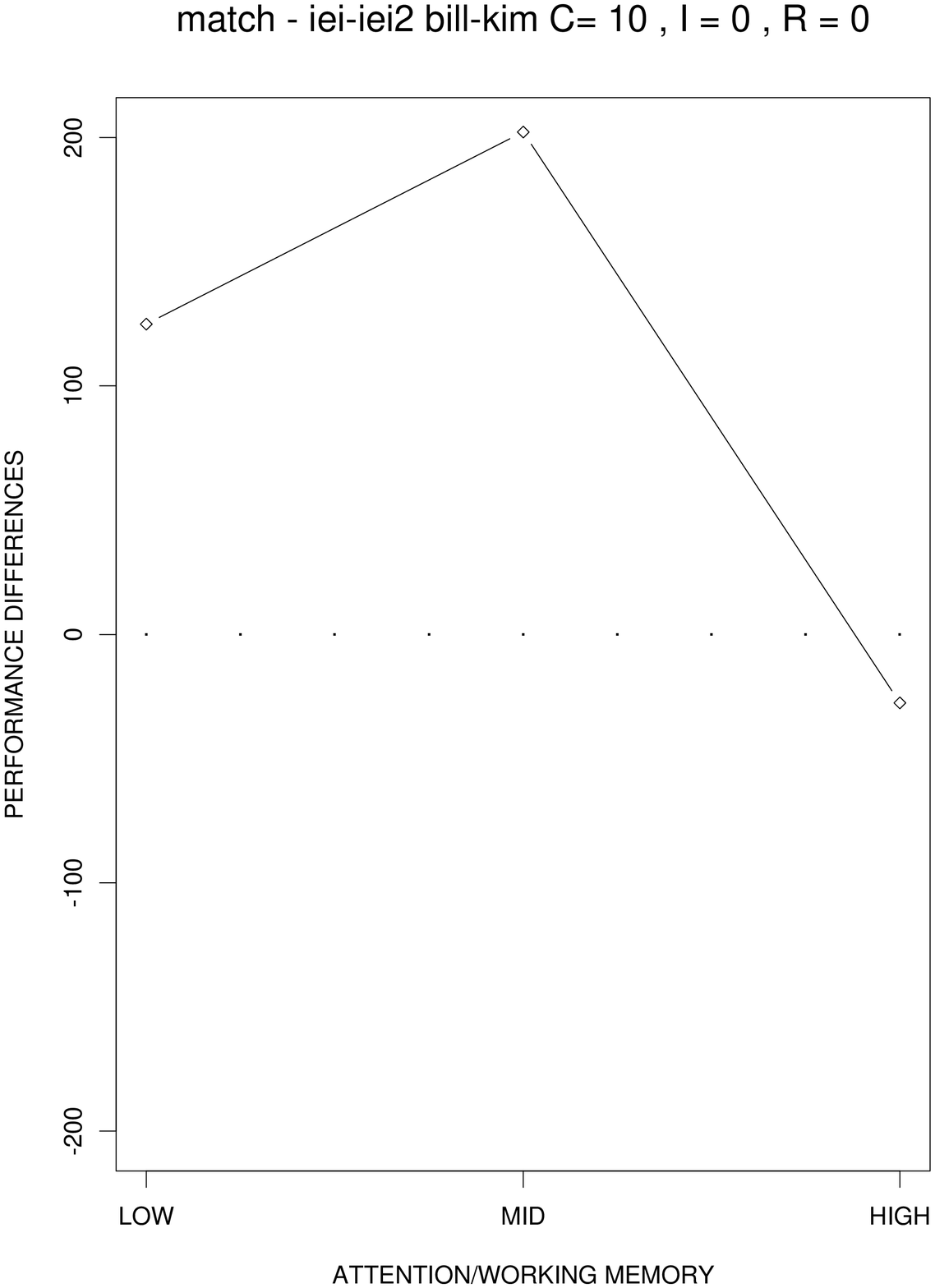,height=3.25in,width=4.5in}}
\caption{Explicit-Warrant is beneficial for Zero-NonMatching-Beliefs
Task, for {\sc low} and {\sc mid awm} agents, even when communication
cost dominates processing: Strategy 1 is two Explicit-Warrant agents
and strategy 2 is two All-Implicit agents: Task =
Zero-Nonmatching-Beliefs, commcost = 10, infcost = 0, retcost = 0}
\label{iei-exp-nmb-fig}
\end{figure}

\subsection{Zero NonMatching Beliefs Task }

Remember that the Zero-Nonmatching-Beliefs task requires a greater
degree of belief coordination by requiring agents to agree on the
beliefs underlying deliberation ({\sc warrants}).\footnote{Remember
that in other tasks, agents do not have to agree on {\sc warrants}
because in situations in which they know of only one option, they do
not need to retrieve the warrant in order to be able to decide to
accept the proposal. Thus when agents have limited {\sc awm}, they may
accept a proposal without having retrieved the warrant.} Thus, it it
increases the importance of making particular deliberation-based
inferences, and can therefore be used to test hypotheses A1, A4 and
A5. Below we will compare the performance of agents using the
All-Implicit strategy with the Explicit-Warrant strategy in the
Zero-NonMatching-Beliefs task.

Figure \ref{iei-nmb-fig} plots the mean performance differences of the
Explicit-Warrant strategy and the All-Implicit strategy in the
Zero-NonMatching-Beliefs task.  A two-way anova exploring the effect
of {\sc awm} and communicative strategy for the Zero-NonMatching-Beliefs
task, shows that {\sc awm} has a large effect on performance (F= 471.42,
p$<$ .000001).  There is also a main effect for communicative strategy
(F = 379.74, p $<$ 0.000001), and an interaction between {\sc awm} and
communicative choice (F= 669.24, p$<$ .000001).

Comparisons within each {\sc awm} range of the two communicative strategies
in this task shows that the Explicit-Warrant strategy is highly
beneficial for {\sc low} and {\sc mid awm} agents ({\sc mb(low)} = 260.6, p
$<$ 0.002; {\sc mb(mid)} = 195.5, p $<$ 0.002). The strategy is also
beneficial for {\sc high awm} agents {\sc mb(high)} = 4.48, p $<$ 0.05).
When agents are resource limited, they may fail to access a warrant.
The Explicit-Warrant strategy guarantees that the agents always can
access the warrant for the option under discussion. Thus,
even agents with higher values of {\sc awm} can benefit from this strategy,
since the task requires such a high degree of belief coordination.

Hypothesis I1 can also be tested in this task. We can ask whether it
is possible to drive the total effort for communication high enough to
make it inefficient to choose the Explicit-Warrant strategy over
All-Implicit. However, the benefits of the Explicit-Warrant strategy
for {\sc low} and {\sc mid awm} agents for this task are so strong
that they cannot be reduced even when communication cost is high
({\sc mb(low)} = 246.4, p $<$ 0.002; {\sc mb(mid)} = 242.7, p $<$ 0.002).
See figure \ref{iei-exp-nmb-fig}.  In other words, even when every
extra {\sc warrant} message increases collaborative effort by 10 and
reduces performance by 10, if the task is Zero-NonMatching-Beliefs,
resource-limited agents using Explicit-Warrant do better.  Contrast
figure \ref{iei-exp-nmb-fig} with the Standard task and same cost
parameters in figure \ref{exp-iei-fig}.

However, when communication cost is high, the strategy becomes
detrimental for {\sc high awm} agents ({\sc mb(high)} = 7.56, p $<$ 0.01).
These agents can usually access warrants and the increase in belief
coordination afforded by the Explicit-Warrant strategy does not offset
the high communication cost.

\begin{figure}[htb]
\centerline{\psfig{figure=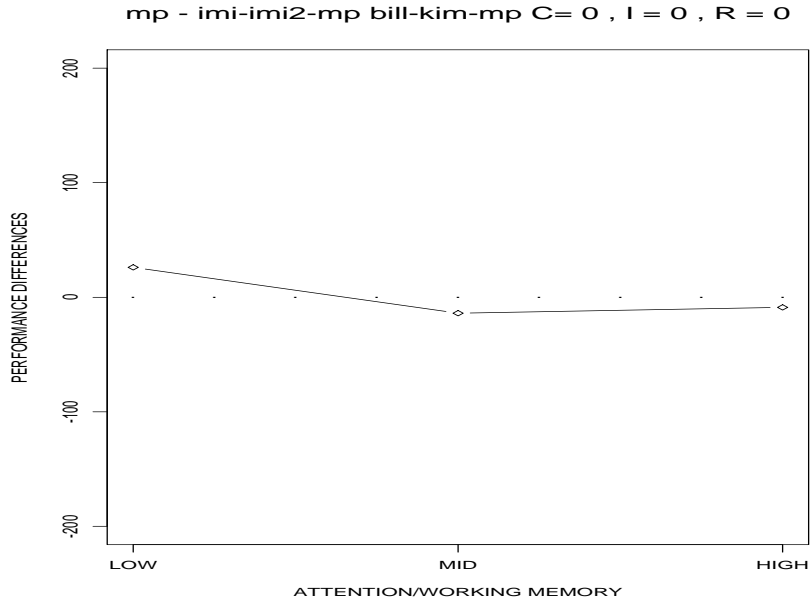,height=3.25in,width=4.5in}}
\caption{Matched-Pair-Inference-Explicit is beneficial for {\sc low
awm}
agents in Matched-Pair-Same-Room. Strategy 1 is two
Matched-Pair-Inference-Explicit agents
and Strategy 2 is two All-Implicit agents,
Task $=$ Matched-Pair-Same-Room,
commcost $=$ 0, infcost $=$ 0, retcost $=$ 0}
\label{imi-imi2-mpr-fig0}
\end{figure}

\begin{figure}[htb]
\centerline{\psfig{figure=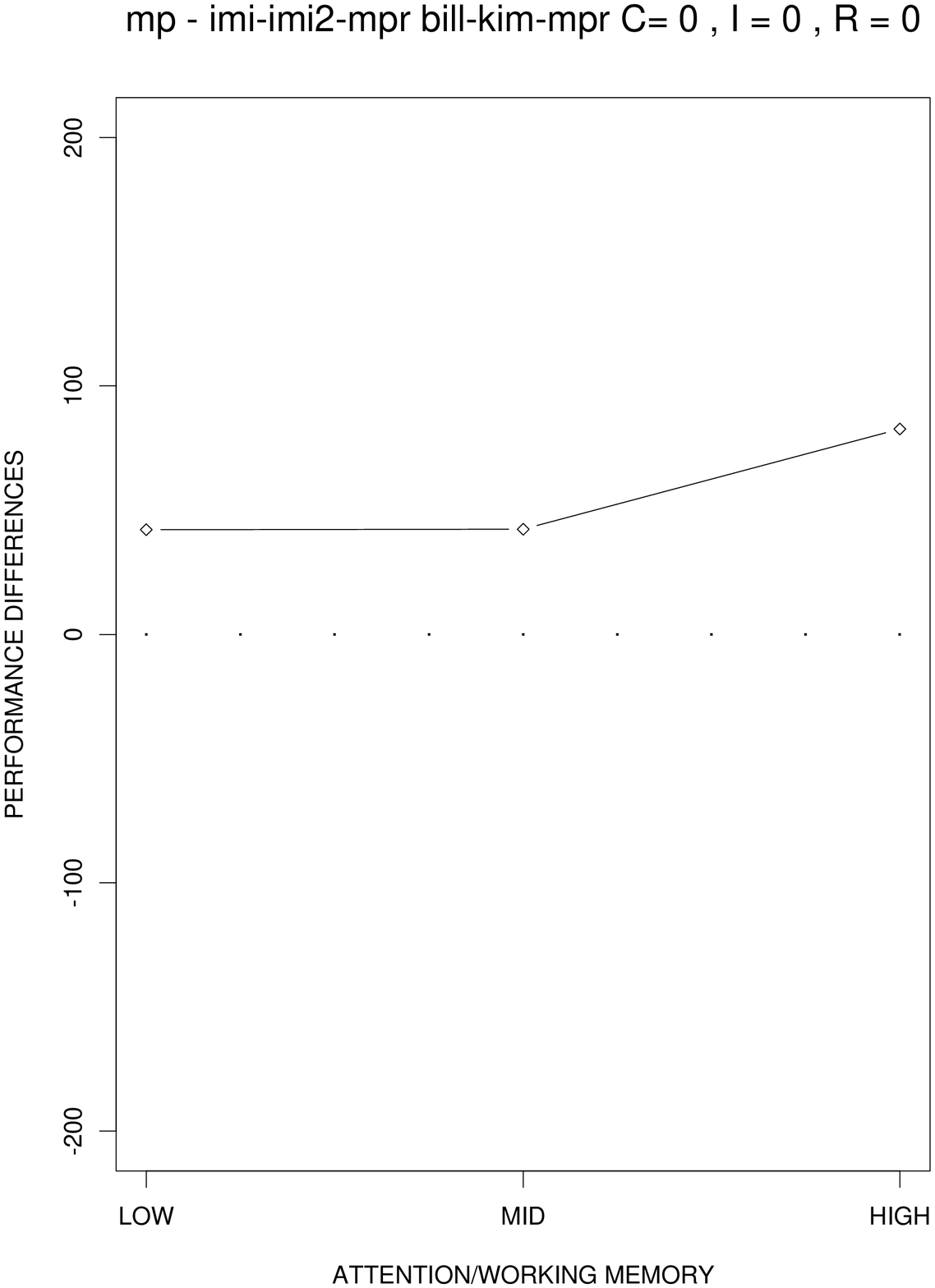,height=3.25in,width=4.5in}}
\caption{Matched-Pair-Inference-Explicit is beneficial  for {\sc low},
{\sc mid} and {\sc high awm} agents in the
Matched-Pair-Two-Room Task. Strategy 1 is two Matched-Pair-Inference-Explicit
agents
and Strategy 2 is two All-Implicit agents,
Task $=$ Matched-Pair-Two-Room,
commcost $=$ 0, infcost $=$ 0, retcost $=$ 0}
\label{imi-imi2-mpr-fig1}
\end{figure}

\begin{figure}[htb]
\centerline{\psfig{figure=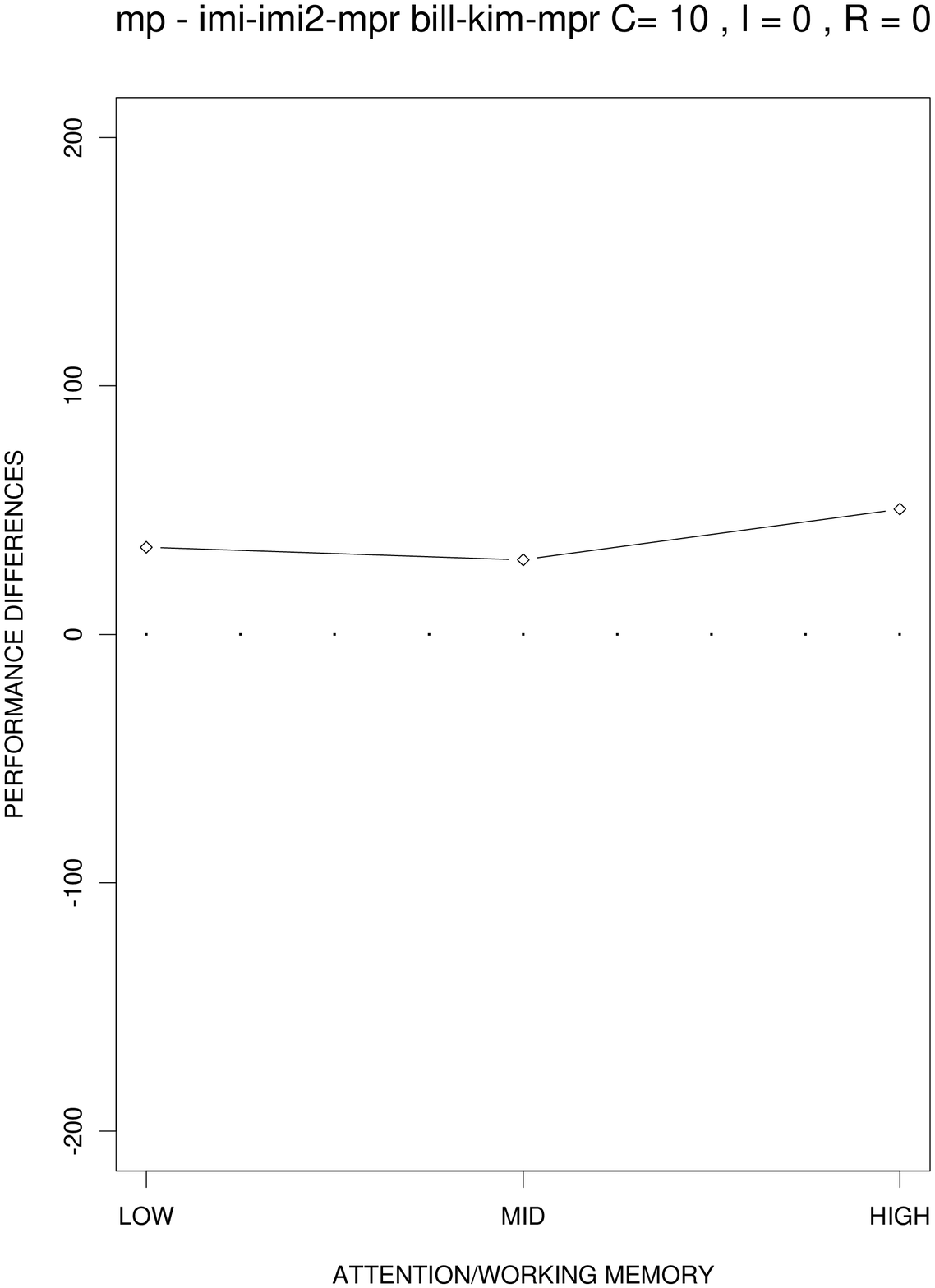,height=3.25in,width=4.5in}}
\caption{The Matched-Pair-Inference-Explicit strategy is beneficial for {\sc
low},
{\sc mid} and {\sc high awm} agents in the
Matched-Pair-Two-Room Task even with communication cost of 10. Strategy 1
is two Matched-Pair-Inference-Explicit agents and Strategy 2 is two
All-Implicit agents, Task $=$ Matched-Pair-Two-Room, commcost $=$ 10,
infcost $=$ 0, retcost $=$ 0}

\label{imi-imi2-mpr-fig4}
\end{figure}

\subsection{Inferential Tasks:   Matched pairs}


The two versions of the Matched-Pair tasks described in section
\ref{task-def-sec} (1) increase the inferential complexity of the task
and (2) increase the degree of belief coordination required by
requiring agents to be coordinated on inferences that follow from
intentions that have been explicitly agreed upon. Both tasks increases
inferential difficulty to a small degree: All-Implicit agents do
fairly well at making matched pair inferences.  The
Matched-Pair-Same-Room task requires the same inferences as the
Matched-Pair-Two-Room task, but these inferences should be easier to
make in the Matched-Pair-Same-Room since the inferential premises are
more likely to be salient.

The Matched-Pair tasks provide an environment for testing hypotheses
A2, A3, A4 and A5.  The Attention strategy that is used to test these
hypotheses is the Matched-Pair-Inference-Explicit strategy; this
strategy makes the premises for matched-pair inferences salient, thus
increasing the likelihood of agents making matched-pair inferences.
The predictions are that this strategy should be beneficial for {\sc
low} and possibly for {\sc mid awm} agents, but that {\sc high awm}
agents can access the necessary inferential premises without Attention
IRUs. Furthermore, we predict that the beneficial effect should be
stronger for the Matched-Pair-Two-Room task.

Figure \ref{imi-imi2-mpr-fig0} plots the performance differences
between All-Implicit agents and Matched-Pair-Inference-Explicit agents
for the Matched-Pair-Same-Room task. A two-way anova exploring the
effect of {\sc awm} and communicative strategy in this task, shows that {\sc
awm}
has a large effect on performance (F= 323.93, p$<$ .000001).  There is
no main effect for communicative strategy (F = .03, ns), but there is
an interaction between {\sc awm} and communicative choice (F= 1101.51, p$<$
.000001).

Comparisons within {\sc awm} ranges between agents using the All-Implicit
strategy and agents using the Matched-Pair-Inference-Explicit strategy
in the Matched-Pair-Same-Room task (figure
\ref{imi-imi2-mpr-fig0}) shows that Matched-Pair-Inference-Explicit
strategy is beneficial for {\sc low awm} agents ({\sc mb(low)} = 4.47, p $<$
.05), but not significantly different for either {\sc mid} or {\sc high
awm} agents). In the Matched-Pair-Same-Room task the content of the
IRU was recently inferred and is likely to still be salient, thus the
beneficial effect is relatively small and is restricted to very
resource limited agents.


In contrast, in the Matched-Pair-Two-Room task, the effect on
performance of the Matched-Pair-Inference-Explicit strategy is much
larger, as we predicted.  Figure \ref{imi-imi2-mpr-fig1} plots the
mean performance differences of agents using the
Matched-Pair-Inference-Explicit strategy and those using the
All-Implicit strategy. The All-Implicit agents do not manage to
achieve the same levels of mutual inference as
Matched-Pair-Inference-Explicit agents.  A two-way anova exploring the
effect of {\sc awm} and communicative strategy in this task, shows that {\sc
awm}
has a large effect on performance (F= 171.79, p$<$ .000001).  There is
a main effect for communicative strategy (F = 57.12, p $<$ .001), and
an interaction between {\sc awm} and communicative choice (F= 567.34, p$<$
.000001).

Comparisons within {\sc awm} ranges between agents using the All-Implicit
strategy and agents using the Matched-Pair-Inference-Explicit strategy
in the Matched-Pair-Two-Room task (figure
\ref{imi-imi2-mpr-fig1}) shows that Matched-Pair-Inference-Explicit
strategy is beneficial for {\sc low, mid} and {\sc high awm} agents
({\sc mb(low)} = 21.94, p $<$ .01); {\sc mb(mid)} = 7.71, p $<$ .01);
{\sc mb(high)} = 38.85, p $<$ .002). In other words, this strategy is highly
effective in increasing the ability of {\sc low}, {\sc mid} and {\sc
high awm} agents to make matched pair inferences in the
Matched-Pair-Two-Room task.

We predicted the strategy to be beneficial for {\sc low} and possibly
for {\sc mid awm} agents because it gives agents access to premises
for inferences which they would otherwise be unable to access. This
confirms the effect of the hypothesized {\sc discourse inference
constraint}.  However, we did not expect it to be beneficial for {\sc
high awm} agents. This surprising effect is due to the fact that, in
the case of higher {\sc awm} values, the Matched-Pair-Inference-Explicit
strategy keeps the agents coordinated on which inference the proposing
agent intended in a situation in which multiple inferences are
possible.  In other words, when agents have {\sc high awm} they can
make {\bf divergent} inferences, and a strategy of making inferential
premises salient improves agents' inferential coordination. Thus the
strategy controls inferential processes in a way that was not
predicted based on the corpus analysis alone.

Hypothesis I1 can also be tested in this task. We can ask whether it
is possible to drive the effort for communication high enough to make
it inefficient to choose the Matched-Pair-Inference-Explicit strategy
over All-Implicit.

Figure \ref{imi-imi2-mpr-fig4} plots the mean performance differences
between these two strategies when communication cost is high.
Comparisons within each {\sc awm} range shows that this strategy is still
beneficial for {\sc low, mid} and {\sc high awm} agents even with a
high communication cost ({\sc mb(low)} = 19.10, p $<$ .01); {\sc mb(mid)} =
3.94,
p $<$ .05); {\sc mb(high)} = 10.46, p $<$ .01).  In other words it would be
difficult to find a task situation that required coordinating on
inference in which this strategy was not beneficial.  This result is
strong support for the {\sc discourse inference constraint}, which may
explain the prevalence of this strategy in naturally occurring
dialogues  \cite{Sadock78,WJ82,RCohen87},

\subsection{Zero Invalids Task }

Remember that the Zero-Invalids Task is a fault-intolerant version of
the task in which any invalid intention invalidates the whole plan.
Thus the Zero-Invalids task provides an environment for testing
hypotheses C2 and C4 with respect to the inferences made explicit by the
Close-Consequence strategy.

\begin{figure}[htb]
\centerline{\psfig{figure=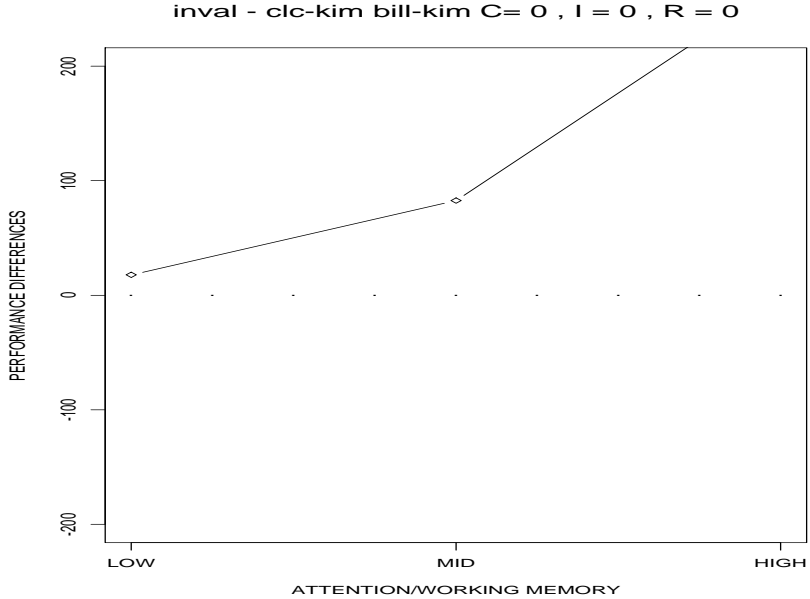,height=3.25in,width=4.5in}}
\caption{Close Consequence is beneficial for the Zero-Invalids Task for {\sc
mid} and {\sc high awm} agents. Strategy 1 is
the combination of an All-Implicit agent with a Close-Consequence
agent and Strategy 2 is two All-Implicit agents, Task = Zero-Invalid,
commcost = 0, infcost = 0, retcost = 0}
\label{clc-inval-fig}
\end{figure}

Figure \ref{clc-inval-fig} plots the mean performance differences
between agents using the Close-Consequence strategy and agents using
the All-Implicit strategy in the Zero-Invalids task.  A two-way anova
exploring the effect of {\sc awm} and communicative strategy in this task,
shows that {\sc awm} has a large effect on performance (F= 223.14, p$<$
.000001).  There is a main effect for communicative strategy (F =
75.81, p $<$ .001), and an interaction between {\sc awm} and communicative
choice (F= 103.38, p$<$ .000001).

The Close-Consequence strategy was detrimental in the Standard task
for {\sc low awm} agents.  Comparisons within {\sc awm} ranges between
agents using the All-Implicit strategy and agents using the
Close-Consequence strategy in the Zero-Invalids task shows that there
are no differences in performance for {\sc low awm} agents in the
fault-intolerant Zero-Invalids task ({\sc mb(low)} = 3.64, ns). However, the
Close-Consequence strategy is beneficial for {\sc mid} and {\sc high awm}
agents ({\sc mb(mid)} = 26.62, p $<$ .002); {\sc mb(high)} = 267.72, p $<$
.002). In other words, this strategy is highly beneficial in
increasing the robustness of the planning process by decreasing the
frequency with which agents make mistakes. This is a direct result of
{\bf rehearsing} the act-effect inferences, making it unlikely that
attention-limited agents will forget these important inferences.

\section{Discussion}
\label{discussion-sec}

This paper showed how agents' choice in communicative action can be
designed to mitigate the effect of their resource limits in the
context of particular features of a collaborative planning task.
In section \ref{model-sec}, I presented a  model of
collaborative planning in dialogue and discussed a number of
parameters that can affect either the efficacy of the final plan or
the efficiency of the collaborative planning process. Then in section
\ref{results-sec}, I presented the results of experiments testing
hypotheses about the effects of these parameters on collaborative
planning dialogues.  These results contribute to the development of
the model of collaborative planning dialogue presented here.  In
addition, since the testbed implementation is compatible with many
current theories, these results could be easily incorporated into
other dialogue planning algorithms
\cite{LRS94,Traum94,Guinn94,GS90,LCN90,groskrau93,ChuCarberry95}, {\it
inter alia}.

A secondary goal of this paper was to argue for a particular
methodology for dialogue theory development.  The method was specified
in section \ref{method-sec}. The Design-World testbed was introduced
in section \ref{dw-sec} and sections \ref{task-def-sec} and
\ref{comm-choice-sec} described the parameterizations of the model
that support testing the hypotheses.  Four parameters for
communicative strategies were tested: (1) All-Implicit; (2)
Close-Consequence; (3) Explicit-Warrant; and (4)
Matched-Pair-Inference-Explicit. Four parameters for tasks were
tested: (1) Standard; (2) Zero-Nonmatching-Beliefs; (3) Matched-Pair
(MP); (4) Zero-Invalid. Three situations of varying processing effort
were tested.

In this section, I will first summarize the hypotheses and the
experimental results in section \ref{summary-sec}, then I will discuss
how the experimental results might generalize to situations not
implemented in the testbed. Section \ref{future-work-sec} proposes
future work and section \ref{conc-sec} consists of concluding
remarks.

\subsection{Summary of Results}
\label{summary-sec}
The hypotheses that were generated by the statistical analysis of the
dialogue corpora are repeated below for convenience from sections
\ref{iru-sec} and \ref{dw-plan-eval-sec}.

\begin{itemize}

\item HYPOTH-C1:  Agents produce Consequence IRUs to demonstrate that they made
the inference that is made explicit.
\item HYPOTH-C2: Agents choose to produce Consequence IRUs to ensure
that
the  other
agent has access to inferrable information.
\item HYPOTH-C3:  The choice to produce a Consequence IRU  is directly related
to a  measure of `how hard' the inference is.
\item HYPOTH-C4: The choice to produce a Consequence IRU  is directly related
to a  measure of `how important' the inference is.
\item HYPOTH-C5: The choice to produce a Consequence IRU  is directly related
to   the degree to which the task requires agents to be coordinated on
the inferences that they have made.

\item HYPOTH-A1:  Agents produce Attention IRUs to support the
processes of deliberating beliefs and intentions.

\item HYPOTH-A2: There is a {\sc discourse inference constraint}
whose effect is that inferences in dialogue are
derived from propositions that are currently discourse salient (in
working memory).

\item HYPOTH-A3: The choice to produce an Attention IRU  is related
to   the degree  of inferential complexity of a task as measured by
the number of premises required to make task related inferences.
\item HYPOTH-A4: The choice to produce an Attention IRU  is related
to the degree to which an agent is resource limited in attentional
capacity.
\item HYPOTH-A5: The choice to produce an Attention IRU  is related
to the degree to which the task requires agents to be coordinated on
the inferences that they have made.
\item HYPOTH-I1: Strategies that reduce collaborative effort without
affecting quality of solution
are beneficial.

\end{itemize}

Below I will summarize the experimental results  reported in
section \ref{results-sec} with respect to the hypotheses above.

Hypotheses C3 and C4 were tested by comparing the Close-Consequence
strategy with the All-Implicit strategy in the Standard task.  In this
experimental setup, the inference made explicit by the Consequence IRU
was neither hard to make nor critical for performance.  Hypothesis C3
was only weakly tested by the experiments because agents always make
this inference.  The results in figure \ref{cost-clc-kim-fig} show
that the Close-Consequence strategy is detrimental for {\sc low awm}
agents. This is because IRUs can displace useful information from
working memory and because the inference made explicit with this IRU
is not `hard enough'.

The Standard task also provides a weak test of hypothesis C4.  The
fact that the Standard task is fault tolerant means that making the
inference is not as critical as it might be.  However, errors can
results from either not making the inference or forgetting it once it
is made.  At lower values of {\sc awm}, the probability of such errors is
not that high.  However, the results shown in figure
\ref{cost-clc-kim-fig} show that the probability of error is
 higher for {\sc high awm} agents in this case, because of their
belief deliberation algorithm, and thus the Close-consequence strategy
is beneficial for {\sc high awm} agents, even in the Standard task.

The Zero-Invalids task provides another test of hypothesis C4 by
increasing the importance of the inference made explicit by the
Close-consequence strategy.  Figure \ref{clc-inval-fig} shows that
hypothesis C4 is confirmed because the Close-Consequence strategy is
beneficial for {\sc low, mid} and {\sc high awm} agents. In addition
to the reasons discussed for the Standard task, this strategy is
beneficial for {\sc high awm} agents because they have more potential
to improve their scores by ensuring that they don't make errors.

The experiments did not test hypothesis C1 because agents in the
testbed are not designed to actively monitor evidence from other
agents as to what inferences they might have made.  Hypothesis C5 was
not tested by the experiments because agents always rectify the
situation if they detect a discrepancy in beliefs about act effect
inferences: they reject proposals whose preconditions do not hold.

Hypotheses A1, A4 and A5 were tested by experiments in which the
Explicit-Warrant strategy was compared with the All-Implicit strategy
in the Standard task.  Hypothesis A1 is disconfirmed for {\sc low awm}
agents. Figure \ref{free-ret-iei-fig} shows that the Explicit-Warrant
strategy is neither beneficial nor detrimental for {\sc low awm}
agents for the Standard task, when processing is free.  This
counterintuitive result arises because, when agents are highly
resource limited, IRUs can displace other information that is more
useful.

To test hypothesis I1 in this situation, we also examined two
situations where processing is not free. When communication cost
dominates other processing costs, the Explicit Warrant strategy is
detrimental for {\sc low} and {\sc high awm} agents. However, when
retrieval cost dominates other processing costs, the Explicit Warrant
strategy is beneficial for {\sc mid awm} agents and there is a trend
toward a beneficial effect for {\sc high awm} agents. Thus these two
situations show that hypothesis I1 is confirmed: processing effort has
a major effect on whether a strategy is beneficial.

We also tested hypotheses A1, A4 and A5 with experiments in which the
Explicit-Warrant strategy was compared with the All-Implicit strategy
in the Zero-Nonmatching-Beliefs task (see figures \ref{iei-nmb-fig}
and \ref{iei-exp-nmb-fig}). This task increases the
importance of making deliberation based inferences by requiring agents
to be coordinated on these inferences in order to do well on the task.
In this situation, we saw a very large beneficial effect for the
Explicit-Warrant strategy, which is not diminished by increasing
communication effort. Thus in situations in which agents are required
to be coordinated on these inferences, strategies which include
Attention IRUs can be very important.

Hypotheses A2, A3, A4, and A5 were tested by experiments comparing the
Matched-Pair-Inference-Explicit strategy with the All-Implicit
strategy in the two versions of the Matched-Pair task.  The results
shown in figures \ref{imi-imi2-mpr-fig0} and \ref{imi-imi2-mpr-fig1}
provide support for these hypotheses.  However these results also
included an unpredicted benefit of Attention IRUs for inferentially
complex tasks where agents must coordinate on inferences. Figure
\ref{imi-imi2-mpr-fig1} shows that both {\sc mid} and {\sc high awm}
agents' performance improves with the Matched-Pair-Inference-Explicit
strategy.  This can be explained by the fact that Attention IRUs
increase the likelihood that agents will make the {\bf same}
inference,
rather than {\bf divergent} inferences,
when multiple inferences are possible.

Furthermore, although the Matched-Pair-Inference-Explicit strategy is
specifically tied to Matched-Pair inferences, it provides a test of a
general strategy for making premises for inferences salient, when
tasks are inferentially complex and require agents to remain
coordinated on inferences. Thus it provides strong support for the
{\sc discourse inference constraint}. To generalize this strategy to
other cases of plan-related inferences, the clauses in the strategy
plan operator that specifically refer to matched-pair inferences can
be replaced with a more general inference, e.g. the more general
(Generates ?act1 $\wedge$ ?act2 ?act3), where the generates relation
is to be inferred \cite{Pollack86a,GS90,Dieugenio93}.

Hypothesis I1 was tested by examining extremes in cost ratios for
retrieval effort and communication effort whenever a hypothesis about
the beneficial effects of IRUs was confirmed. Figure \ref{exp-iei-fig}
shows that high communication effort can make the Explicit-Warrant
strategy detrimental in the Standard task.  Figure
\ref{iei-exp-nmb-fig} shows that high communication effort does not
eliminate the benefits of the Explicit-Warrant strategy in the
Zero-Nonmatching-Beliefs task.  Figure
and \ref{imi-imi2-mpr-fig4} shows that high communication effort does
not eliminate the benefits of the Matched-Pair-Inference-Explicit
strategy in the Matched-Pair-Two-Room task.  Thus the strategy of
making premises for inferences salient is robust against extremes in
processing effort.

\subsection{Generalizability of the Results}

This section addresses concerns raised in \cite{HPC93} that simulation
is ` experimentation in the small'. Hanks writes that (\cite{HPC93},
section 5.1.5):

\begin{quote}

The ultimate value-- arguably the {\em only}  value -- of
experimentation is to constrain or otherwise inform the designer of a system
that solves interesting  problems.  In order to do so the experimenter must
demonstrate three things:
\begin{enumerate}
\item that her results -- the relationships she demonstrates between agent
characteristics and world characteristics -- extend beyond the particular
agent, world, and problem specification she studied,
\item that the solution to the problem area she studied in isolation will be
applicable when that same problem area is encountered in a larger, more complex
world, and
\item that the relationship demonstrated experimentally actually constrains or
somehow guides the design of a larger more realistic agent.
\end{enumerate}
\end{quote}

The list in 1 to 3 are all different ways of saying that the results
should generalize beyond the specifics of the experiment, and this
after all is a basic issue with all experimental work.  Typically
generalizations can be shown by a series of multiple experiments
modifying multiple variables as we have done here.  For example, the
modifications to the task are specifically designed to test whether
beneficial communicative strategies generalize across tasks.  However,
we might also ask to what extent do the variables manipulated in the
simulation abstract out key properties of real situations?  Below I
will briefly discuss why the results presented above are potentially
generalizable. I will focus on generalizations along three dimensions:
(1) task (or environmental) properties; (2) agent architectural
properties; and (3) agent behaviors. These dimensions are the same as
those in Cohen's `ecological triangle' \cite{Cohen89}.

\paragraph{Generalizations about tasks}

The Design-World task was selected as a simple planning task that
requires negotiation of each step. The structure of this task is
isomorphic to a subcomponent of many collaborative planning tasks.  In
addition, to test generalizability of hypothesized benefits across
tasks, we examined more complex variants of the task by manipulating
three abstract features: (1) inferential complexity as measured by the
number of premises required for making a task related inference and
(2) degree of belief coordination required on intentions, inferences
and beliefs underlying a plan; and (3) the task determinacy and fault
tolerance of the plan.  These general features can certainly be
applied to other tasks in other domains. In fact it is difficult to
think of a task or domain in which these features could not be
applied.

\paragraph{Generalizations about agent  properties}

Design-World agents are artificial agents that are designed to model
the resource limited qualities of human agents.  The planning and
deliberation aspects of human processing are modeled with the IRMA
architecture, and resource limits on these processes are modeled by
extending the IRMA architecture with a model of Attention/Working
Memory ({\sc awm}) which has been shown to model a limited but critical set
of properties of human processing.  The way that agents process
dialogue is tied to the agent architecture.

The experimental results will extend to dialogues between artificial
agents to the extent that those agents exhibit similar cognitive
properties.  Here, we looked at a resource bound on access to memory
as modeled by a size of memory subset limit, however size is directly
correlated to {\bf time} to access memory. Artificial agents are often
time limited in rapidly changing worlds, so it seems quite plausible
that artificial agents would benefit from similar communicative
strategies.  For example, I would predict that agents in the Phoenix
simulation testbed would benefit from the strategies discussed here
\cite{Cohen89}.  In other work artificial agents do `make inferences
explicit' by communicating to other agents partial computations when
the other agent might have been able to make these computations
\cite{Cohen89,DLC87,Turner94}.  In addition, defining inferential
complexity as a direct consequence of the number of premises simultaneously in
memory
bears a strong resemblance to problems artificial processors have when
a computation requires a large working set \cite{Stone87}.

The experimental results should extend to dialogues between humans and
artificial agents because Design-World agents are designed to model
humans. However it may be desirable to change the definition of
collaborative effort for modeling human-computer interaction to allow
the computer to handle processing that is easy for the computer to do
and for the human to handle processing that is easy for the human to
do. Furthermore, most of the claims about the {\sc awm} model are based on a
limited set of human working memory properties, and these properties
will also hold for other cognitively based architectures such as SOAR
\cite{Soar87,LehmanSoar}.

\paragraph{Generalizations about agent behaviors}

In this work the agent behaviors that were tested were the agent
communication strategies.  One reason to believe that the strategies
are general to human-human discourse is that they were based on
observed strategies in different corpora of natural collaborative
planning dialogues.  It is possible to find all three types of IRUs in
the Trains, Map-Task and Design corpora
\cite{trains,Carletta92,PHW82,WGR93},  as well as in the financial
advice domain.

In addition to this empirical evidence, there are further reasons why
we might expect generalizations.

The communicative acts and discourse acts used by Design-World agents
are similar to those used in
\cite{Carletta92,Cawsey92a,Sidner94,Stein93}.  Thus communicative
strategies based on these acts should be implementable in any of these
systems.

The experimental results based on these strategies should generalize
to other discourse situations because the strategies are based on
general relations between utterance acts and underlying processes,
such as supporting deliberation and inference.  For example, the
mapping of a {\sc warrant} relation between an act and a belief in
naturally occurring examples such as \ref{walnut-examp} was modeled
with a {\sc warrant} relation between an act and a belief in
Design-World, as seen in the Explicit-Warrant communication strategy.
The claims made about the use of the Explicit-Warrant communication
strategy
should generalize to any dialogue planning domain where agents
use warrants to support deliberation.

Similarly, content based inferences in natural dialogues such as that
discussed in relation to example \ref{certif-examp} were modeled with
content based inferences in Design-World such as those required for
doing well on the Matched-Pair tasks. This inferential situation was
designed to test the {\sc discourse inference constraint}, that
inferences in dialogue are restricted to premises that are currently
salient. Both experimental and corpus based evidence was provided in
support of the discourse inference constraint. The claims made about
the use of the Matched-Pair-Inference-Explicit communication
strategy, based on experimental evidence, should generalize to any
dialogue strategy where agents make premises for inferences available,
and to any planning domain where agents are required to make content
based inferences in support of deliberation or planning.

The evaluation metrics applied to these strategies should also
generalize whenever domain plan utility is a reasonable measure of the
quality of solution for a dialogue task.

\subsection{Relation to Other Work}


The model of collaborative planning dialogues presented in section
\ref{model-sec} draws from previous work on cooperative dialogue
\cite{WJ82,PHW82,Litman85,Pollack86a,JWW86,GS86,FJW86,Carberry89,CS89,WS88,GS90,RCohen87}, and the results are applicable to other current research on
collaborative planning
\cite{Sidner94,CohenLevesque91,HeemanHirst95,ChuCarberry95,Guinn94,Traum94,Dahlback91,Lochbaum,groskrau93,YMP94}.

The agent architecture and the model of deliberation and means-end
reasoning is based on the work of \cite{BIP88} and \cite{Doyle92}, and
on Pollack's TileWorld simulation environment
\cite{PollackRinguette90}. The use of IRMA as an underlying model
of intention deliberation to provide a basis for a collaborative
planning model was first proposed in
\cite{Walker92a,Walker93e,Walker93c}, and has been incorporated into
other work \cite{groskrau93,YMP94}.  The architecture includes a
specific model of limited working memory, but most of the claims about
the model are based on its recency and frequency properties, which
might also be provided by other cognitively based architectures such
as SOAR\cite{Soar87,LehmanSoar}. \footnote{\cite{Walker95b} discusses
the differences between an {\sc awm}-like attentional model and Grosz
and Sidner's stack model of attentional state
\cite{GS86,Sidner79,Grosz77}. See also \cite{Rose95} for a discussion
of other discourse phenomena for which the {\sc awm} model may be
useful.} Since the testbed architecture is consistent with that
assumed in other work, the experimental results should be
generalizable to those frameworks.

The relationship between discourse acts and domain-based options and
intentions in this work is based on Litman's model of discourse plans
\cite{Litman85,LA90} and is similar to the approach in
\cite{Carletta92,Cawsey92a,Traum94}.  The emphasis on autonomy at each stage of
the
planning process and the belief reasoning mechanism of Design-World
agents is based on the theory of belief revision and the
multi-agent simulation environment developed in the Automated
Librarian project \cite{Galliers89,Galliers91b,Cawsey92a,LRS94}.

The Design-World testbed is based on the methods used in the TileWorld
and Phoenix simulation environments: rapidly changing robot worlds in
which an artificial agent attempts to optimize reasoning and planning
\cite{PollackRinguette90,HPC93,Cohen89}.  TileWorld is a single agent
world in which the agent interacts with its environment, rather than
with another agent.  Design-World uses similar methods to test a
theory of the effect of resource limits on communicative behavior
between two agents.

Design-World is also based on the method used in Carletta's JAM
simulation for the Edinburgh Map-Task \cite{Power74,Carletta92}.  JAM
is based on the Map-Task Dialogue corpus, where the goal of the task
is for the planning agent, the instructor, to instruct the reactive
agent, the instructee, how to get from one place to another on the
map.  JAM focuses on efficient strategies for recovery from error and
parametrizes agents according to their communicative and error
recovery strategies.  Given good error recovery strategies, Carletta
argues that `high risk' communicative strategies are more efficient,
but did not attempt to quantify efficiency.  In contrast, the approach
here provides a way of quantifying what is an effective or efficient
strategy, and the results suggest that a combination of the agents'
resource limitations and the task definition determine when strategies
are efficient. Future work could test Carletta's claims about recovery
strategies within this extended framework.

To my knowledge, none of this earlier work has considered the factors
that affect the range of variation in communicative choice, or the
effects of different choices, or measured how communicative choice
affects the construction of a collaborative plan and the ability of
the conversants to stay coordinated.  Nor have other theories of
collaborative planning been explicit about the agent architecture, or
tested specific ideas about resource bounds in dialogue, and none have
used utility as the basis for agents' communicative choice.  In
addition, no earlier work on cooperative task-oriented dialogue argued
that conversational agents' resource limits and task complexity are
major factors in determining effective conversational strategies in
collaboration.

\subsection{Future Work}
\label{future-work-sec}
A promising avenue for future work is to investigate beneficial
strategies for teams of heterogeneous agents. In the experiments here,
pairs of agents in dialogue were always parameterized with the same
resource limits. Pilot studies of dialogues between heterogeneous
agents suggest that strategies that are not effective for homogeneous
agents may be effective for heterogeneous ones. For example, in
\cite{Walker93c} I tested an Attention IRU strategy in which agents
would tell one another about all the options they knew about at the
beginning of planning each room. This strategy is not beneficial for
homogeneous agents because IRUs can displace other useful information.
However if one agent is not limited, then it can be helpful for the
resource limited agent to exploit the capabilities of the more
capable agent by telling the other agent important facts before it
forgets them.

Another extension would be to extend the agent communication
strategies or to test additional ones. For example, other work
proposes a number of strategies for information selection and ordering
in dialogue and provides some evidence that these strategies are
efficient or efficacious
\cite{CareniniMoore93,Suthers93,ZukermanMcConachy93,ChuCarberry95}.
Support for these claims could be provided by Design-World experiments
in which agents used these strategies to communicate.

Future work could also modify the properties of the world or of the
task.  For example, it would be possible to make Design-World more
like Tileworld by making the world change in the course of the task,
by adding or removing furniture.

These results may also be incorporated as input into decision
algorithms in which agents decide online which strategy to pursue, and
investigate additional factors that determine when strategies are
effective in collaborative planning dialogues.  The results presented
here show what information an agent should consider. For example, a
comparison between {\sc low, mid} and {\sc high awm} agents shows how
to design decision algorithms for agents who have to decide whether to
expend additional effort.

Another promising avenue is make the agents capable of remembering and
learning from past mistakes so that they can adapt their strategies to
the situation \cite{AZC91}.

Finally, these results should be incorporated into the design of
multi-agent problem-solving systems and into systems for
human-computer communication, such as those for teaching, advice and
explanation, where for example the use of particular strategies might
be premised on the abilities of the learner or apprentice.

\subsection{Concluding Remarks}
\label{conc-sec}
The goal of this paper was to show how agents' choice in communicative
action, their algorithms for language behavior, can be designed to
mitigate the effect of their resource limits in the context of
particular features of a collaborative planning task.  In this paper,
I first motivate a number of hypotheses based on a statistical
analysis of natural collaborative planning dialogues.  Then a
functional model of collaborative planning dialogues is developed
based on these hypotheses, including parameters that are hypothesized
to affect the generalizability of the model. The model is then
implemented in a testbed in which these parameters can be varied, and
the hypotheses are tested.

The method used here can be contrasted with other work on dialogue
modeling. Much previous work on dialogue modeling only carries out
part of the process described above: only the initial part of the
process up to specifying a functional model is completed. Followon
research that is based on these models must judge the model according
to subjective criteria such as how well it fits researcher's
intuitions or how elegant the model is. The models developed here on
the basis of empirical evidence can also be judged according to these
subjective criteria, but this work carries out additional steps to
further test and refine the model suggested by the corpus analysis.
Implementing a model with parameters to test the generalizability of
the model and testing hypotheses in a testbed implementation provides
a way to check  subjective evaluations and suggests many ways in
which our initial hypotheses must be refined and further tested.

The Design-World testbed is the first testbed for conversational
systems that systematically introduces several different types of
independent parameters that are hypothesized to affect the efficacy of
a collaborative plan negotiated through a dialogue, and the efficiency
of that dialogue process.  Experiments in the testbed examined the
interaction between (1) agents' resource limits in attentional
capacity and inferential capacity; (2) agents' choice in
communication; and (3) features of communicative tasks that affect
task difficulty such as inferential complexity, degree of belief
coordination required, and tolerance for errors.  The results verified
a number of hypotheses that depended on particular assumptions about
agents' resource limits that were not possible to test by corpus
analysis alone.

Several unpredicted and counterintuitive results were also
demonstrated by the experiments.  First, the task property of belief
coordination in {\bf combination} with resource limits (as in the
Zero-Nonmatching-Beliefs and Matched-Pair tasks), were shown to
produce the most robust benefits for IRUs, rather than resource limits
alone as originally hypothesized.  Second, I predicted that IRUs would
always be beneficial for {\sc low awm} agents, but found that IRUs can
be detrimental for these agents through a side effect of displacing
other, more useful, beliefs from working memory.  Third, it would seem
plausible that {\sc high awm} agents should always perform better than
either {\sc low} or {\sc mid awm} agents since these agents always
have access to more information.  However the results showed that
there are two situations in which this is not an advantage: (1) when
accessing information has some cost; and (2) when access to multiple
beliefs can lead agents to make divergent inferences.  In this case,
restricting agents to a small shared working set is a natural way to
limit inferential processes.  This limit intuitively corresponds to
potential benefits of limited working memory for humans and explains
how humans manage to coordinate on inferences in conversation
\cite{Levinson85,Grosz77,Joshi78}.

These results clearly demonstrate that factors not previously
considered in dialogue models must be taken into account of claims if
cooperativity, efficiency, or efficacy are to be supported. In
addition, I have shown that a theory of dialogue that includes a model
of resource-limited processing can account for both the observed
language behavior in human-human dialogue and the experimental results
presented here.

\nocite{Kidd85,Hobbs85a,CohenLevesque91,Jameson,ChuCarberry94,Rosenschein85,Litman85,Carletta92,GS90,Sidner94,Traum,trains,WS88,WW90,Walker93c}

\section{Acknowledgements}

The work reported in this paper has benefited from discussions with
Steve Whittaker, Aravind Joshi, Ellen Prince, Mark Liberman, Max
Mintz, Bonnie Webber, Scott Weinstein, Candy Sidner, Owen Rambow, Beth
Ann Hockey, Karen Sparck Jones, Julia Galliers, Phil Stenton, Megan
Moser, Johanna Moore, Christine Nakatani, Penni Sibun, Ellen Germain,
Janet Cahn, Jean Carletta, Jon Oberlander, Julia Hirschberg, Alison
Cawsey, Rich Thomason, Cynthia McLemore, Jerry Hobbs, Pam
Jordan, Barbara Di Eugenio, Susan Brennan, Rebecca Passonneau, Rick
Alterman and Paul Cohen. I am grateful to Julia Galliers for providing
me with an early implementation of the belief revision mechanism used
in the Automated Librarian project, and to Julia Hirschberg who
provided me with tapes of the financial advice talk show. Thanks also
to the two anonymous reviewers who provided many useful suggestions.

This research was partially funded by ARO grants DAAG29-84-K-0061 and
DAAL03-89-C0031PRI, DARPA grants N00014-85-K0018 and N00014-90-J-1863,
NSF grants MCS-82-19196 and IRI 90-16592 and Fellowship INT-9110856
for the 1991 Summer Science and Engineering Institute in Japan, and
Ben Franklin Grant 91S.3078C-1 at the University of Pennsylvania, and
by Hewlett-Packard Laboratories.


\end{document}